\documentclass[conference]{IEEEtran}
\IEEEoverridecommandlockouts
\usepackage{cite}
\usepackage{amsmath,amssymb,amsfonts}
\usepackage{algorithmic}
\usepackage{graphicx}
\usepackage{textcomp}
\usepackage[table,xcdraw]{xcolor}
\usepackage{xcolor}
\usepackage{hhline}
\usepackage{enumitem}
\usepackage{stfloats}
\usepackage[footnotesize]{subfig}
\usepackage{multirow}
\def\BibTeX{{\rm B\kern-.05em{\sc i\kern-.025em b}\kern-.08em T\kern-.1667em\lower.7ex\hbox{E}\kern-.125emX}}
\newcommand{\scheme}{MODEF}
\newcommand*\rot{\rotatebox{90}}
\makeatletter

\newcommand*{\centerfloat}{%
\parindent \z@
\leftskip \z@ \@plus 1fil \@minus \textwidth
\rightskip\leftskip
\parfillskip \z@skip}
\makeatother
\begin{document}
	
\title{Denoising and Verification Cross-Layer Ensemble \\
	Against Black-box Adversarial Attacks} 

\author{
\IEEEauthorblockN{Ka-Ho Chow, Wenqi Wei, Yanzhao Wu, Ling Liu}
\IEEEauthorblockA{
	School of Computer Science \\
	Georgia Institute of Technology\\
	Atlanta, GA, USA 30332}
}
\maketitle

\begin{abstract}
Deep neural networks (DNNs) have demonstrated impressive performance on many challenging machine learning tasks. However, DNNs are vulnerable to adversarial inputs generated by adding maliciously crafted perturbations to the benign inputs. As a growing number of attacks have been reported to generate adversarial inputs of varying sophistication, the defense-attack arms race has been accelerated. In this paper, we present \scheme{}, a cross-layer model diversity ensemble framework. \scheme{} intelligently combines unsupervised model denoising ensemble with supervised model verification ensemble by quantifying model diversity, aiming to boost the robustness of the target model against adversarial examples. Evaluated using eleven representative attacks on popular benchmark datasets, we show that \scheme{} achieves remarkable defense success rates, compared with existing defense methods, and provides a superior capability of repairing adversarial inputs and making correct predictions with high accuracy in the presence of black-box attacks. 
\end{abstract}
\begin{IEEEkeywords}
adversarial deep learning, ensemble defense, ensemble diversity, robustness
\end{IEEEkeywords}

\vspace{-1.1em}
\section{Introduction}
The recent advances in deep neural networks (DNNs) have powered numerous applications in different domains due to their outstanding performance compared to traditional machine learning techniques. However, it has been shown that DNNs can be easily fooled by adversarial inputs~\cite{goodfellow2014explaining} and become a double-edged sword as their vulnerability to adversarial attacks has posed serious threats to many security-critical applications, such as biometric authentication and autonomous driving. As a number of defenses are being proposed, more attacks of varying sophistication have been put forward,  accelerating the defense-attack arms race. Some even argue that designing new attacks requires much less efforts than developing effective defenses. Thus, improving the robustness and defensibility against adversarial attacks is crucial.

Adversarial examples are generated by maliciously perturbing benign examples sent to the target DNN model through querying its prediction API, aiming to fool and mislead the target model to misclassify by producing incorrect predictions randomly (untargeted attack) or purposefully (targeted attack). Given a $D$-dimensional benign example $\boldsymbol{x}\in\mathbb{R}^D$ and a $K$-class classification target model $\text{TM}:\mathbb{R}^D\rightarrow\mathbb{R}^K$ such that the prediction is given as $C_{\boldsymbol{x}}=\arg\max_{1\leq i\leq K}\text{TM}_i(\boldsymbol{x})$, the generation process of the adversarial example $\boldsymbol{x}'$ can be formulated as
\vspace{-2mm}
\begin{equation}\small\label{eq:adv-perturb}
\min||\boldsymbol{x} - \boldsymbol{x}'||_{p} \quad s.t.\;C_{\boldsymbol{x}'} = {y^*},\ C_{\boldsymbol{x}'} \ne C_{\boldsymbol{x}},
\end{equation}
where  $p$ is the distance metric and $y^{*}$ denotes the target class label for targeted attacks and any incorrect class label for untargeted attacks. For image inputs, the distance metric $p$ can be $0$, $2$, or $\infty$ representing $L_0$, $L_2$, and $L_\infty$ norms where $L_0$ norm counts the number of pixels of $\boldsymbol{x}$ that are changed, $L_2$ norm is the Euclidean distance between $\boldsymbol{x}$ and $\boldsymbol{x}'$, and $L_\infty$ norm denotes the maximum change to any pixel of $\boldsymbol{x}$. 

{\bf Related Work.\/} Existing defense proposals are mostly attack-dependent and can be broadly divided into three categories. {\em Adversarial training} counters known attacks by retraining the target model using adversarial examples generated by each attack algorithm~\cite{madry2017towards, kurakin2016adversarial,tramer2017ensemble}. {\em Input transformation} defenses apply noise reduction techniques to input data to reduce the sensitivity of the target model to small changes due to adversarial perturbations. However, to distinguish adversarial examples from benign ones, they rely on finding a dataset-specific or attack-specific threshold~\cite{meng2017magnet,xu2017feature, samangouei2018defense}. {\em Gradient masking} defenses aim to harden the process of generating adversarial examples by hiding the gradient information from attackers, such as distillation training techniques~\cite{papernot2016distillation}. But they either significantly reduce the benign accuracy of the target model or are vulnerable to attack transferability~\cite{papernot2017practical,carlini2017towards}. 

In this paper, we present \scheme{}, a cross-layer MOdel Diversity Ensemble Framework by integrating unsupervised model denoising ensemble with supervised model verification ensemble, aiming to protect the target model against adversarial attacks. From a manifold learning perspective~\cite{chapelle2009semi}, natural high-dimensional data concentrate close to a nonlinear low-dimensional manifold, and adversarial attacks can be considered as a malicious process to drag benign examples away from the manifold where they concentrate. \scheme{} first exploits denoising autoencoders to learn such manifold and to map adversarial examples back to their corresponding benign form. Different from prior works using denoising autoencoders as an input preprocessing-based defense~\cite{sahay2019combatting, liao2018defense}, \scheme{} leverages multiple denoisers to boost defensibility by joint force and quantifies model diversity to enable strategic teaming of denoisers with diverse denoising effects. To further improve the robustness and repair those adversarial examples escaped from the denoising autoencoders, the denoised example is sent to our second layer of model verification ensemble, which exploits the weak spots of attack transferability using a team of failure-independent models, for prediction verification and repairing. We further enhance the defensibility of \scheme{} with defense structure randomization by creating a pool of models and enabling strategic randomized ensemble teaming. By promoting such uncertainty at runtime, it allows \scheme{} to combat adversarial attacks with higher robustness.

In summary, \scheme{} advances existing works from three perspectives. First, it  provides three model diversity ensemble defenses against adversarial examples: the model denoising ensemble, the model verification ensemble, and the denoising-verification cross-layer ensemble, each improves the previous layer by empowering the target model with higher robustness. Second, \scheme{} is by design attack-independent and can generalize well over attack algorithms. \scheme{} does not use dataset-specific or attack-specific magic parameters, such as the detection thresholds, to distinguish adversarial examples from benign ones. Furthermore, \scheme{} enables strategic defense structure randomization to improve its defensibility against adversarial examples and hardens black-box attacks~\cite{papernot2017practical}. Extensive experiments on popular benchmark datasets with eleven representative attacks are used to validate that \scheme{} can significantly improve the robustness of a target DNN model against adversarial attacks.

The remainder of this paper is organized as follows. We first briefly review the representative attacks and the benchmark datasets used in this paper in Section~\ref{sec:attacks}. Then, we introduce the Model Denoising Ensemble Defense in Section~\ref{sec:dnn-denoising} and the Model Verification Ensemble Defense and the Denoising-Verification Cross-Layer Ensemble Defense in Section~\ref{sec:co-defense}, followed by experimental evaluation in Section~\ref{sec:experimental-results}. We conclude the paper in  Section~\ref{sec:conclusion}.

\vspace{-2mm}
\section{Adversarial Examples}\label{sec:attacks}
Adversarial examples are generated over their corresponding benign examples by adding maliciously crafted perturbations that can fool the target model to misclassify during the model prediction (testing) phase. For the attack threat model, if the adversary can generate adversarial examples by only accessing the prediction API of the target model with no knowledge of the target model training process, we call such attack the black-box attack. In this paper, we focus on defense strategies against black-box attacks.

We generate adversarial examples using seven representative attack algorithms: FGSM~\cite{goodfellow2014explaining}, BIM~\cite{kurakin2016adversarial}, CW$_{0}$, CW$_{2}$, CW$_{\infty}$~\cite{carlini2017towards}, DeepFool~\cite{moosavi2016deepfool}, and JSMA~\cite{papernot2016limitations} which cover a wide spectrum of techniques including both $L_\infty$, $L_2$, and $L_0$ distortions. We build those attacks on top of EvadeML-Zoo~\cite{xu2017feature} using the same set of hyperparameters. For each targeted attack, we study two attack targets representing two ends of the targeted attack spectrum: the most-likely (ML) attack class in the prediction vector (i.e., $y^* = \arg\max_{1\leq i\leq K,i \ne C_{\boldsymbol{x}}} \text{TM}_i(\boldsymbol{x})$) and the least-likely (LL) attack class (i.e., $y^* = \arg\min_{1\leq i\leq K} \text{TM}_i(\boldsymbol{x})$). Thus we have a total of eleven different attacks.  

Table~\ref{tab:attack-evaluation} summarizes the evaluation of the attacks for two popular benchmark datasets: MNIST and CIFAR-10. MNIST consists of $70,000$ gray-scale images of ten handwritten digits, each image is $28\times28\times1$ in size with $60,000$ images for training and the remaining $10,000$ images for testing. CIFAR-10 consists of $60,000$ colorful images of ten classes, each is $32\times32\times3$ in size. Similarly, for CIFAR-10, $50,000$ images are used for training with the remaining $10,000$ images for testing. For MNIST, the target model is a seven-layer CNN~\cite{carlini2017towards}  with an accuracy of $0.9943$. For CIFAR-10, the target model is a DenseNet~\cite{huang2017densely} with an accuracy of $0.9484$. The first $100$ testing examples ($10$ per class) that are correctly classified by the target model are selected to generate adversarial examples. We exclude DeepFool (DF) from MNIST because the generated images are unrecognizable to humans.
\begin{table}
	\scriptsize
	\caption{Evaluation of the attacks on MNIST and CIFAR-10.}
		\vspace{-1mm}
	\label{tab:attack-evaluation}
	\begin{tabular}{|c|c|c|c|c|c|c|c|c|c|c|}
		\hline
		\multirow{2}{*}{\textbf{}} & \multicolumn{3}{c|}{\textbf{Configuration}} & \multirow{2}{*}{\textbf{Cost (s)}} & \multirow{2}{*}{\textbf{ASR}} & \multirow{2}{*}{\textbf{MR}} & \multirow{2}{*}{\textbf{\begin{tabular}[c]{@{}c@{}}Prediction\\ Confidence\end{tabular}}} & \multicolumn{3}{c|}{\textbf{Distortion}} \\ \cline{2-4} \cline{9-11} 
		& \textbf{} & \textbf{Attack} & \textbf{Mode} &  &  &  &  & \textbf{L$_\infty$} & \textbf{L$_2$} & \textbf{L$_0$} \\ \hline
		\multirow{10}{*}{\rot{MNIST}} & \multirow{4}{*}{$L_\infty$} & FGSM & \multirow{2}{*}{UA} & $0.003$ & $0.46$ & $0.46$ & $0.9475$ & $0.302$ & $5.931$ & $0.563$ \\ \cline{3-3} \cline{5-11} 
		&  & BIM &  & $0.01$ & $0.92$ & $0.92$ & $0.9982$ & $0.302$ & $4.819$ & $0.522$ \\ \cline{3-11} 
		&  & \multirow{2}{*}{CW$_\infty$} & ML & $57.2$ & $1.00$ & $1.00$ & $0.9999$ & $0.226$ & $3.235$ & $0.416$ \\ \cline{4-11} 
		&  &  & LL & $50.1$ & $1.00$ & $1.00$ & $0.9998$ & $0.279$ & $4.655$ & $0.507$ \\ \cline{2-11} 
		& \multirow{2}{*}{$L_2$} & \multirow{2}{*}{CW$_2$} & ML & $0.3$ & $1.00$ & $1.00$ & $0.9999$ & $0.603$ & $2.151$ & $0.443$ \\ \cline{4-11} 
		&  &  & LL & $0.3$ & $1.00$ & $1.00$ & $0.9999$ & $0.733$ & $3.207$ & $0.458$ \\ \cline{2-11} 
		& \multirow{4}{*}{$L_0$} & \multirow{2}{*}{CW$_0$} & ML & $65.1$ & $1.00$ & $1.00$ & $0.9999$ & $0.983$ & $3.667$ & $0.029$ \\ \cline{4-11} 
		&  &  & LL & $61.8$ & $1.00$ & $1.00$ & $0.9999$ & $0.995$ & $5.122$ & $0.061$ \\ \cline{3-11} 
		&  & \multirow{2}{*}{JSMA} & ML & $0.6$ & $0.93$ & $0.93$ & $0.7137$ & $1.000$ & $3.728$ & $0.031$ \\ \cline{4-11} 
		&  &  & LL & $0.7$ & $0.43$ & $0.53$ & $0.6253$ & $1.000$ & $5.555$ & $0.063$ \\ \hhline{|=|=|=|=|=|=|=|=|=|=|=|}
		\multirow{11}{*}{\rot{CIFAR-10}} & \multirow{4}{*}{$L_\infty$} & FGSM & \multirow{2}{*}{UA} & $0.1$ & $0.86$ & $0.86$ & $0.9694$ & $0.016$ & $0.864$ & $0.998$ \\ \cline{3-3} \cline{5-11} 
		&  & BIM &  & $0.4$ & $0.92$ & $0.92$ & $0.9872$ & $0.008$ & $0.367$ & $0.994$ \\ \cline{3-11} 
		&  & \multirow{2}{*}{CW$_\infty$} & ML & $186.0$ & $1.00$ & $1.00$ & $0.9890$ & $0.009$ & $0.341$ & $0.960$ \\ \cline{4-11} 
		&  &  & LL & $197.6$ & $1.00$ & $1.00$ & $0.9751$ & $0.014$ & $0.522$ & $0.994$ \\ \cline{2-11} 
		& \multirow{3}{*}{$L_2$} & DF & UA & $0.5$ & $1.00$ & $1.00$ & $0.8382$ & $0.028$ & $0.028$ & $0.993$ \\ \cline{3-11} 
		&  & \multirow{2}{*}{CW$_2$} & ML & $6.1$ & $1.00$ & $1.00$ & $0.9866$ & $0.024$ & $0.204$ & $0.641$ \\ \cline{4-11} 
		&  &  & LL & $7.0$ & $1.00$ & $1.00$ & $0.9730$ & $0.041$ & $0.353$ & $0.847$ \\ \cline{2-11} 
		& \multirow{4}{*}{$L_0$} & \multirow{2}{*}{CW$_0$} & ML & $459.4$ & $1.00$ & $1.00$ & $0.9898$ & $0.581$ & $1.547$ & $0.010$ \\ \cline{4-11} 
		&  &  & LL & $461.2$ & $1.00$ & $1.00$ & $0.9764$ & $0.694$ & $2.517$ & $0.024$ \\ \cline{3-11} 
		&  & \multirow{2}{*}{JSMA} & ML & $7.5$ & $1.00$ & $1.00$ & $0.5326$ & $0.843$ & $3.681$ & $0.046$ \\ \cline{4-11} 
		&  &  & LL & $12.0$ & $0.98$ & $1.00$ & $0.3984$ & $0.904$ & $5.563$ & $0.098$ \\ \hline
	\end{tabular}
	\vspace{-5.5mm}
\end{table}

Attack success rate (ASR) measures the percentage of successful adversarial examples over all attacked inputs while misclassification rate (MR) is defined as the percentage of misclassified adversarial examples over all attacked inputs. For untargeted attacks (UA), MR is the same as ASR, but for targeted attacks, MR may be higher than ASR because an adversarial example may fail the targeted attack but still cause misclassification, resulting in a wrong prediction output other than the true class. Most adversarial attacks evaluated produce a high ASR and MR with high prediction confidence, meaning that the target model being attacked predicts adversarial examples to be a wrong class with a high probability.

We utilize the following metrics to evaluate defensibility: \emph{Prevention Success Rate (PSR)} measures the percentage of the adversarial examples that are repaired and correctly classified by the target model under defense. With the benign testing set, PSR denotes the benign accuracy. \emph{Detection Success Rate (TSR)} computes the percentage of adversarial examples that could not be repaired but are correctly flagged as the attack example by the defense. For the benign testing set, TSR is the benign false positive rate, i.e., the percentage of the benign examples being flagged as adversarial over the total number of benign examples. \emph{Defense Success Rate (DSR)} is the percentage of adversarial examples that are either repaired or detected ($\text{DSR} = \text{PSR} + \text{TSR}$). \emph{False Positive Rate (FP)} measures the percentage of the adversarial examples that can be correctly classified (repaired) but are flagged as adversarial when all inputs are adversarial examples. For the benign testing set, it represents the percentage of the correctly classified benign examples being flagged as adversarial.
All evaluations on benign examples are based on the entire testing set with $10,000$ images.

\vspace{-1mm}
\section{Model Denoising Ensemble Defense}\label{sec:dnn-denoising}
We design the model denoising ensemble defense as the first perimeter of defense to prevent adversarial misclassification. The defense structure consists of multiple DNN denoisers, each performs noise reduction via unsupervised learning using one specific denoising autoencoder, aiming to remove adversarial perturbations as much as possible by joint force. 

\vspace{-1mm}
\subsection{Denoising Autoencoders}
Denoising autoencoders are initially proposed as a way to extract and compose robust features which can be utilized to replace and optimize the random initialization and bootstrap the efficiency of training deep neural networks~\cite{vincent2008extracting}. From a manifold learning perspective~\cite{chapelle2009semi}, natural high-dimensional data concentrate close to a nonlinear low-dimensional manifold. Adversarial attacks can be considered as a malicious process to drag benign examples away from the manifold where they concentrate. DNN denoising training is to learn a function to map a corrupted example, likely to be outside and farther from the manifold, back to its uncorrupted form. Thus, we can utilize denoising autoencoders, trained with uniformly corrupted examples, to reverse the adversarial perturbation process such that the adverse effect can be removed. Several efforts~\cite{xie2012image,agostinelli2013adaptive} have been proposed to perform image denoising tasks with different DNN models and different design of autoencoder structures and hyperparameters. 
\begin{figure*}  
	\centering
	\includegraphics[width=0.75\linewidth]{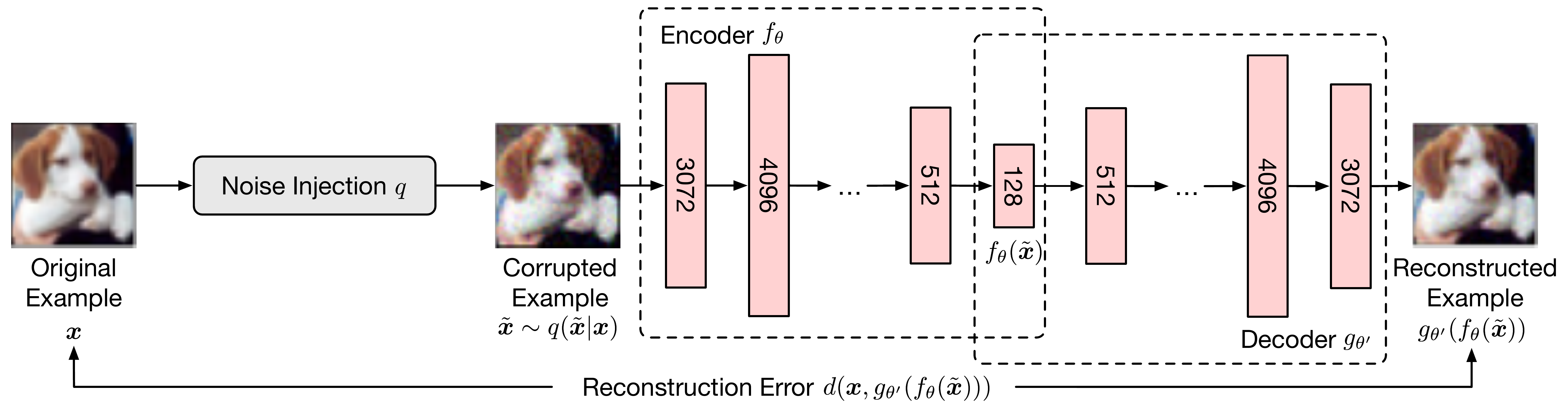}
			\vspace{-2mm}
	\caption{The training process of a denoising autoencoder with an example from CIFAR-10.}
	\label{fig:dae-training}
	\vspace{-5mm}
\end{figure*}
For image data, a denoising autoencoder takes an input image, transforms it into a noisy version, and feeds the noisy image to the autoencoder to perform latent space projection and then reconstructs the image with the goal of generating a clean version of the original image as the output. Figure~\ref{fig:dae-training} illustrates, by example, the key components of training a denoising autoencoder. 

Let $\boldsymbol{x}$ be an example from the training set and $\tilde{\boldsymbol{x}}$ be the version corrupted by a stochastic noise mapping $q$ such that $\tilde{\boldsymbol{x}}\sim q(\tilde{\boldsymbol{x}}\vert\boldsymbol{x})$~\cite{vincent2010stacked}. The encoder is an $L_f$-layer neural network, which projects the corrupted example $\tilde{\boldsymbol{x}}$ from a high-dimensional image space to a low-dimensional latent feature space, producing its latent representation $f_\theta(\tilde{\boldsymbol{x}})=f^{L_f}(\cdots(f^{2}(f^{1}(\tilde{\boldsymbol{x}};\theta_1);\theta_2));\theta_{L_f})$ where $f^i$ is the operation at the $i$-th encoding layer (e.g., convolution) with weights $\theta_i$, $f_\theta$ and $\theta=(\theta_{1},...,\theta_{L_f})$ can be viewed as the composite function of the encoder and the weights respectively. The decoder, an $L_g$-layer neural network, then restores the spatial structure of $f_\theta(\tilde{\boldsymbol{x}})$ by mapping its latent representation back to the original image feature space and produces the reconstructed example $g_{\theta'}(f_\theta(\tilde{\boldsymbol{x}}))=g^{L_g}(\cdots(g^{2}(g^{1}(f_\theta(\tilde{\boldsymbol{x}});\theta'_1);\theta'_2));\theta'_{L_g})$ where $g^i$ is the operation at the $i$-th decoding layer (e.g., deconvolution) with weights $\theta_i'$, $g_{\theta'}$ and $\theta'=(\theta'_{1},...,\theta'_{L_g})$ can be viewed as the composite function of the decoder and the weights respectively. The multilayer encoder and decoder constitute a deep denoising autoencoder. Given $N$ training examples $\{\boldsymbol{x}_1,...,\boldsymbol{x}_N\}$, the denoising autoencoder is trained by backpropagation to minimize the reconstruction loss:
\begin{equation}\small\label{eq:dae-opt}
\begin{split}
& \mathcal{L}(\theta, \theta'; \{\boldsymbol{x}_i\}^{N}_{i=1},d, q, \{f^i\}^{L_f}_{i=1}, \{g^i\}^{L_g}_{i=1},\lambda)\\
=&\frac{1}{N}\sum_{i=1}^{N}d(\boldsymbol{x}_i, g_{\theta'}(f_{\theta}(\tilde{\boldsymbol{x}}_i))) + \frac{\lambda}{2}(\vert\vert\theta\vert\vert^2_\text{F}+\vert\vert\theta'\vert\vert^2_\text{F}),
\end{split}
\end{equation}
where $d$ is a distance function and $\lambda$ is a regularization hyperparameter penalizing the Frobenius norm of $\theta$ and $\theta'$. Given a query example $\boldsymbol{x}$ at runtime, the denoising autoencoder produces a denoised version $\mathcal{D}(\boldsymbol{x})=g_{\theta'}(f_\theta(\boldsymbol{x}))$ with fixed weights $(\theta, \theta')$.

\vspace{-2mm}
\subsection{Strategic Teaming of Multiple DNN Denoisers}
\begin{figure*}
	\centering
	\includegraphics[width=0.75\linewidth]{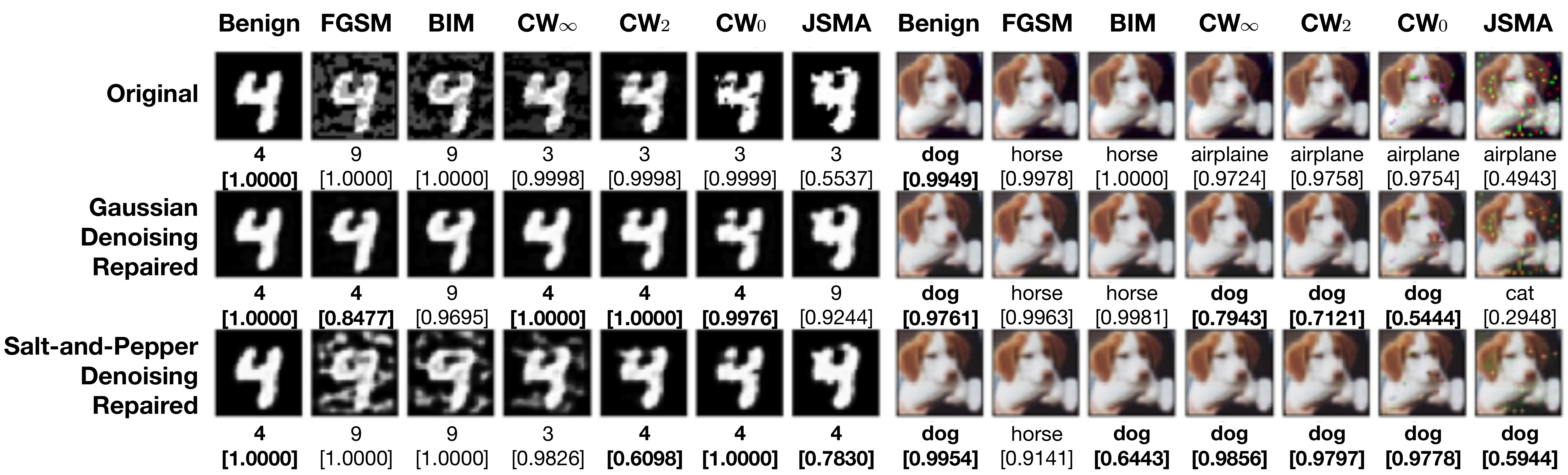}
				\vspace{-1mm}
	\caption{The visualization of denoising effects by two denoising autoencoders on MNIST (left) and CIFAR-10 (right).}
	\label{fig:denoising-example}
	\vspace{-5mm}
\end{figure*}

Recall in Figure~\ref{fig:dae-training} that the noise injection function $q$ of a denoising autoencoder transforms the original input $\boldsymbol{x}$ to a noisy version of $\boldsymbol{x}$, denoted by $\tilde{\boldsymbol{x}}$. Clearly, different data corruption methods produce different versions of input $\boldsymbol{x}$ and result in denoisers that exhibit different denoising effects. Figure~\ref{fig:denoising-example} visualizes two testing examples from MNIST and CIFAR-10 and the corresponding adversarial examples generated by six attack methods (the 1st row), their denoised versions produced by two denoisers trained with Gaussian noise (the 2nd row) and salt-and-pepper noise (the 3rd row) respectively. The predicted class label and confidence by the target model are presented for each case. On the first row, it shows that adversarial attacks successfully fool the target model under all six attacks for the examples from both datasets. On the second row, Gaussian denoiser successfully remove malicious perturbations from four out of six attacks for MNIST example and three out of six attacks for CIFAR-10 example, showing the robustness improvement of the target model even with single DNN denoiser. On the third row, the salt-and-pepper denoiser successfully repaired five out of six attacks for CIFAR-10 example and three out of six attacks for MNIST example. 
These examples deliver two important messages: First, exploiting denoising autoencoders trained with uniformly corrupted examples can remove meticulously crafted adversarial perturbations and restore the classification capability of the target model. Second, no single denoiser is effective across all attacks and each of them is good at removing some types of noise but not the others. This motivates us to employ a team of diverse denoisers, instead of relying on a single denoiser, such that an adversarial example can be denoised from different perspectives.

In our model denoising ensemble defense, we first construct a set of base DNN denoisers by exploiting different approaches to generate denoisers.  For example, we could use different input transformation techniques to perform the input corruption process $q$ (e.g., Gaussian noise, salt-and-pepper noise, and masking noise) such that the optimization in Equation~\ref{eq:dae-opt} attempts to learn different functions to reverse the corruption~\cite{xie2012image}. Alternatively, we can also create different denoisers by altering the model structure (i.e., $\{f^i\}^{L_f}_{i=1}$ and $\{g^i\}^{L_g}_{i=1}$), or the hyperparameter settings (e.g., $\lambda$, training epochs, and random seed for weight initialization), because different ways of generating denoisers can have different effects with respect to the manifold and the convergence of the DNN learning~\cite{wei2018adversarial, huang2017snapshot,wu2018comparative}. The third approach to generate different denoisers is to use different optimization objectives, such as changing the distance function $d$ from a simple per-pixel loss to a perceptual loss~\cite{johnson2016perceptual}, or using an advanced regularization such as a sparsity constraint~\cite{cho2013simple}. 

Upon obtaining a set of base denoisers, we next define ensemble diversity metric to identify and select diverse ensembles such that each of the ensemble members has high benign accuracy and the ensemble teaming has high diversity in terms of failure-independence and low negative example correlation. In our first prototype of \scheme{}, we use the kappa coefficient ($\kappa$)~\cite{mchugh2012interrater} as a pairwise metric to quantify the diversity of an ensemble by measuring each pair of its member denoisers from a statistical perspective. The top-ranked ensemble teams by kappa diversity from the set of base denoisers will be chosen as the pool of our candidate ensembles. 

{\bf Kappa Diversity Metric.\/} Let $N^-$ denote the number of testing examples that the denoised versions are labeled as different classes by the target model, $N^-_{ij}$ denote the number of instances in the testing set that the target model labels one denoised version as class $i$ and the other as class $j$. The kappa metric can be computed by
\begin{equation}\small\label{eq:cohen-kappa}
\kappa=\frac{\frac{1}{N^-}\sum_{i=1}^{K}N^-_{ii}-\sum_{i=1}^{K}{(\frac{N^-_{i*}}{N^-}-\frac{N^-_{*i}}{N^-})}}{1-\sum_{i=1}^{K}(\frac{N^-_{i*}}{N^-}-\frac{N^-_{*i}}{N^-})}.
\end{equation}
Based on the kappa value of each pair of base denoisers, we enumerate and rank all possible ensemble teaming combinations according to their average kappa values and maintain a ``$\kappa$-ranked list'' of denoising ensembles with the low average kappa values below a system-supplied threshold to ensure the ensemble used by \scheme{} has high denoising diversity.

{\bf Denoising Ensemble Selection and Output Decision.\/} We select one denoising ensemble team from the $\kappa$-ranked list of denoising ensembles at runtime. For the chosen denoising ensemble of size $Z_{D}$, we send the query (test) example $\boldsymbol{x}$ to each ensemble member and obtain $Z_{D}$ denoised versions of $\boldsymbol{x}$. There are several ways to produce the ensemble output accordingly. In the first prototype of \scheme{}, we use a majority voting method. When the target model makes a consistent prediction on a majority of the denoised versions, we randomly choose one of the majority members and uses its denoised version as the ensemble output. Otherwise, the query example is flagged as adversarial.

Figure~\ref{fig:denoising-example} provides an illustrative example. First, the benign case refers to the target model without attacks (the 1st column for MNIST and the 8th column for CIRAR-10). The 2nd to the 7th columns correspond to six attacks for MNIST and the 9th to the 14th correspond to six attacks for CIFAR-10. The first row shows that the attacks to the two original test examples (digit $4$ in MNIST and dog in CIFAR-10) are successful with high confidence. However, when we employ the two denoisers (see the 2nd row and the 3rd row), the six adversarial attacks may fail for both test examples with either Gaussian denoiser or salt-and-pepper denoiser. For MNIST digit $4$, the denoising ensemble will survive under CW$_{0}$ and CW$_{2}$. For CIFAR-10 dog example, the denoising ensemble will survive under all three CW attacks. This example also illustrates that some attack algorithms, such as FGSM, BIM and JSMA, can escape from the denoising ensemble defense. One may argue that the reconstruction vectors from a well-trained denoising autoencoder form a vector field which points in the direction of the data manifold. Yet, this may not hold for the examples distant from the manifold as they are rarely sampled during training~\cite{alain2014regularized}. Thus, a more comprehensive solution approach should be developed. This motivates us to propose the output model verification ensemble (Section~\ref{sec:multi-model-verification}) and to integrate the input denoising and output verification to formulate a cross-layer model ensemble defense (Section~{\ref{sec:cross-layer}}).

\vspace{-1mm}
\section{Denoising and Verification Co-Defense}\label{sec:co-defense}
In this section, we first describe the model verification ensemble as an alternative defense method and then present our denoising-verification cross-layer ensemble defense framework, which takes the model denoising ensemble as the front-end defense and then combines it with the model verification ensemble as the back-end defense. 
\vspace{-1mm}
\subsection{Model Verification Ensemble Defense}\label{sec:multi-model-verification}
One of the intriguing properties of adversarial examples is the attack transferability~\cite{papernot2016transferability}. The main idea of model verification ensemble is to break adversarial attacks by exploiting the weak spots of adversarial transferability. It is known that two diverse models trained on the same dataset for the same learning task, their decision boundaries can be quite different even if they obtain a similar training accuracy. Hence, an adversarial example generated by one attack algorithm through one way of perturbing a benign input may successfully fool the target model but may not succeed in fooling other models trained on the same dataset, especially when such models are diverse with respect to negative examples and thus are failure-independent. By creating model ensembles with high diversity as output verifiers for the target model, we argue that (1) the model verification ensembles can improve the robustness of the target model against adversarial examples, and (2) the model verification ensembles are complementary and alternative to denoising ensembles, and thus a cross-layer diversity ensemble defense that integrates input denoising ensemble with output verification ensemble provides greater potential for higher robustness, because the adversarial examples are unlikely to be transferable to all of the member models of the cross-layer ensemble.  

Similar to the creation of denoising ensembles, we perform two steps to create verification ensembles. First, we construct a set of base models as the verifiers for examining and repairing the target model prediction output, each is trained on the same dataset for the same task as the target model. Several techniques can be used to produce the base candidate models, such as varying neural network structures~\cite{wu2018comparative}, training hyperparameters, or performing data augmentation~\cite{krizhevsky2012imagenet}. One can also conduct snapshot learning~\cite{huang2017snapshot} to obtain a set of model verifiers efficiently in a single run of model training. We employ two pruning criteria to generate model diversity ensembles. The first filter is to use the test accuracy to select only those base candidate models with the test accuracy similar to that of the target model. For MNIST and CIFAR-10, we collect numerous pre-trained DNN models from the public domain, with high training and test accuracy under no attack scenarios. The second filter is to select those model ensemble teams that have high model diversity with respect to high failure-independence and low negative/error correlation. In the first prototype of \scheme{}, we utilize the kappa coefficient metric to quantify the disagreement between each pair of base candidate models.  Let $N^-$ be the number of testing examples the base candidate models produce inconsistent predictions and $N^-_{ij}$ be the number of instances in the testing set, which are labeled as class $i$ by one model and class $j$ by the other in Equation~\ref{eq:cohen-kappa}. We then enumerate and rank all possible ensemble teaming combinations with a minimum team size of $3$ and we compute the average kappa value for each ensemble team examined and select a $\kappa$-ranked list of model verification ensembles with low average kappa values using the system-defined kappa-diversity threshold. Any model ensemble from this chosen $\kappa$-ranked list will have its average kappa value below the given threshold and thus high model diversity.

\begin{table}
	\scriptsize
	\centering
	\caption{The ten model verifiers for each dataset with their prediction accuracy on benign examples.}
	\vspace{-1mm}
	\label{tab:a-verifier-acc}
	\begin{tabular}{|c|c|c|c|c|}
		\hline
		\multirow{2}{*}{\textbf{Model}} & \multicolumn{2}{c|}{\textbf{MNIST}} & \multicolumn{2}{c|}{\textbf{CIFAR-10}} \\ \cline{2-5} 
		& \textbf{Name} & \textbf{Benign Accuracy} & \textbf{Name} & \textbf{Benign Accuracy} \\ \hline
		V$_1$ & CNN-5 & $0.9919$ & CNN-10 & $0.9062$ \\ \hline
		V$_2$ & CNN-4 & $0.9861$ & ResNet20 & $0.9205$ \\ \hline
		V$_3$ & CNN-6 & $0.9917$ & ResNet32 & $0.9313$ \\ \hline
		V$_4$ & LeNet & $0.9880$ & ResNet50 & $0.9312$ \\ \hline
		V$_5$ & MLP & $0.9761$ & ResNet56 & $0.9419$ \\ \hline
		V$_6$ & MobileNet & $0.9934$ & ResNet110 & $0.9419$ \\ \hline
		V$_7$ & MobileNet-v2 & $0.9939$ & ResNet152 & $0.9392$ \\ \hline
		V$_8$ & PNASNet & $0.9950$ & ResNext29 & $0.9725$ \\ \hline
		V$_9$ & SVM & $0.9832$ & VGG & $0.9359$ \\ \hline
		V$_{10}$ & VGG & $0.9946$ & WideResNet28 & $0.9712$ \\ \hline
	\end{tabular}
\vspace{-5mm}
\end{table}

Table~\ref{tab:a-verifier-acc} gives the set of ten base models for MNIST and CIFAR-10 respectively, and all ten models have similar benign accuracy under no attack scenarios. Table~\ref{fig:verification-example} illustrates the model verification ensemble method using the same two query examples: digit $4$ from MNIST and a color image of dog from CIFAR-10. The 2nd row shows how the TM (target model) responds to the query example under benign (no attack) and under the six different attacks for MNIST (columns 2-8) and for CIFAR-10 (columns 9-14). When no defense protection to the target model, all adversarial examples successfully fool the target model for both digit $4$ of MNIST and dog image of CIFAR-10. Interestingly, even with the adversarial example transferability, some of these adversarial examples are correctly classified by some of the ten base model verifiers (denoted by $\text{V}_1,...,\text{V}_{10}$).

\begin{table*}
	\centering
	\scriptsize
	\caption{Predicted class label and confidence of each verifier on benign or adversarial examples. The last two rows show the predictions by model verification ensemble (Section~\ref{sec:multi-model-verification}) and denoising-verification cross-layer ensemble (Section~\ref{sec:cross-layer}).}
	\vspace{-1mm}
	\label{fig:verification-example}
	\begin{tabular}{|r|c|c|c|c|c|c|c|c|c|c|c|c|c|c|}
		\hline
		\multirow{3}{*}{} & \multicolumn{7}{c|}{\textbf{MNIST}} & \multicolumn{7}{c|}{\textbf{CIFAR-10}} \\ \cline{2-15} 
		& \textbf{Benign} & \textbf{FGSM} & \textbf{BIM} & \textbf{CW$_\infty$} & \textbf{CW$_2$} & \textbf{CW$_0$} & \textbf{JSMA} & \textbf{Benign} & \textbf{FGSM} & \textbf{BIM} & \textbf{CW$_\infty$} & \textbf{CW$_2$} & \textbf{CW$_0$} & \textbf{JSMA} \\ 
		& \includegraphics[scale=0.5]{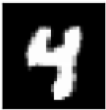} 
		& \includegraphics[scale=0.5]{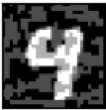} 
		& \includegraphics[scale=0.5]{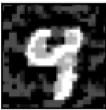} 
		& \includegraphics[scale=0.5]{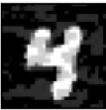}
		& \includegraphics[scale=0.5]{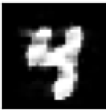}
		& \includegraphics[scale=0.5]{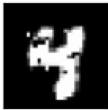}  
		& \includegraphics[scale=0.5]{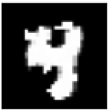}  
		& \includegraphics[scale=0.5]{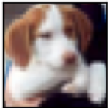}  
		& \includegraphics[scale=0.5]{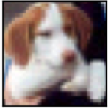} 
		& \includegraphics[scale=0.5]{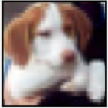}  
		& \includegraphics[scale=0.5]{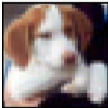}  
		& \includegraphics[scale=0.5]{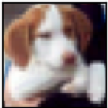}  
		& \includegraphics[scale=0.5]{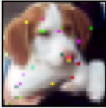}  
		& \includegraphics[scale=0.5]{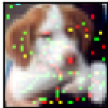}  \\ \hline
		\textbf{TM} & \begin{tabular}[c]{@{}c@{}}$\textbf{4}$\\$\textbf{[1.0000]}$\end{tabular} &  \begin{tabular}[c]{@{}c@{}}$\text 9$\\$\text{[1.0000]}$\end{tabular} &  \begin{tabular}[c]{@{}c@{}}$\text{9}$\\$\text{[1.0000]}$\end{tabular} & \begin{tabular}[c]{@{}c@{}}$\text{3}$\\$\text{[0.9998]}$\end{tabular} & \begin{tabular}[c]{@{}c@{}}$\text{3}$\\$\text{[0.9998]}$\end{tabular} & \begin{tabular}[c]{@{}c@{}}$\text{3}$\\$\text{[0.9999]}$\end{tabular} & \begin{tabular}[c]{@{}c@{}}$\text{3}$\\$\text{[0.5537]}$\end{tabular} & \begin{tabular}[c]{@{}c@{}}\textbf{dog}\\$\textbf{[0.9949]}$\end{tabular} & \begin{tabular}[c]{@{}c@{}}horse\\{[0.9978]}\end{tabular} & \begin{tabular}[c]{@{}c@{}}horse\\{[1.0000]}\end{tabular}  & \begin{tabular}[c]{@{}c@{}}airplane\\{[0.9724]}\end{tabular}  & \begin{tabular}[c]{@{}c@{}}airplane\\{[0.9758]}\end{tabular}  & \begin{tabular}[c]{@{}c@{}}airplane\\{[0.9754]}\end{tabular}  & \begin{tabular}[c]{@{}c@{}}airplane\\{[0.4943]}\end{tabular}  \\ \hline
		\textbf{$\text{V}_1$} & \begin{tabular}[c]{@{}c@{}}$\textbf{4}$\\$\textbf{[0.9029]}$\end{tabular} & \begin{tabular}[c]{@{}c@{}}9\\{[0.3287]}\end{tabular} & \begin{tabular}[c]{@{}c@{}}{9}\\{[0.6249]}\end{tabular} & \begin{tabular}[c]{@{}c@{}}{9}\\{[0.2235]}\end{tabular} & \begin{tabular}[c]{@{}c@{}}$\textbf{4}$\\$\textbf{[0.6314]}$\end{tabular} & \begin{tabular}[c]{@{}c@{}}$\textbf{4}$\\$\textbf{[0.7399]}$\end{tabular} & \begin{tabular}[c]{@{}c@{}}{9}\\{[0.5120]}\end{tabular} & \begin{tabular}[c]{@{}c@{}}$\textbf{dog}$\\$\textbf{[0.9530]}$\end{tabular} & \begin{tabular}[c]{@{}c@{}}$\textbf{dog}$\\$\textbf{[0.4945]}$\end{tabular} & \begin{tabular}[c]{@{}c@{}}$\textbf{dog}$\\$\textbf{[0.7772]}$\end{tabular} & \begin{tabular}[c]{@{}c@{}}$\textbf{dog}$\\$\textbf{[0.8809]}$\end{tabular} & \begin{tabular}[c]{@{}c@{}}$\textbf{dog}$\\$\textbf{[0.7491]}$\end{tabular} & \begin{tabular}[c]{@{}c@{}}$\textbf{dog}$\\$\textbf{[0.9023]}$\end{tabular} & \begin{tabular}[c]{@{}c@{}}{truck}\\{[0.8392]}\end{tabular} \\ \hline
		\textbf{$\text{V}_2$} & \begin{tabular}[c]{@{}c@{}}$\textbf{4}$\\$\textbf{[1.0000]}$\end{tabular} & \begin{tabular}[c]{@{}c@{}}9\\{[1.0000]}\end{tabular} & \begin{tabular}[c]{@{}c@{}}{9}\\{[1.0000]}\end{tabular} & \begin{tabular}[c]{@{}c@{}}$\textbf{4}$\\$\textbf{[0.9925]}$\end{tabular} & \begin{tabular}[c]{@{}c@{}}$\textbf{4}$\\$\textbf{[1.0000]}$\end{tabular} & \begin{tabular}[c]{@{}c@{}}$\textbf{4}$\\$\textbf{[0.9997]}$\end{tabular} & \begin{tabular}[c]{@{}c@{}}$\textbf{4}$\\$\textbf{[0.9934]}$\end{tabular} & \begin{tabular}[c]{@{}c@{}}$\textbf{dog}$\\$\textbf{[1.0000]}$\end{tabular} & \begin{tabular}[c]{@{}c@{}}$\textbf{dog}$\\$\textbf{[1.0000]}$\end{tabular} &\begin{tabular}[c]{@{}c@{}}$\textbf{dog}$\\$\textbf{[1.0000]}$\end{tabular}  & \begin{tabular}[c]{@{}c@{}}$\textbf{dog}$\\$\textbf{[1.0000]}$\end{tabular} &\begin{tabular}[c]{@{}c@{}}$\textbf{dog}$\\$\textbf{[1.0000]}$\end{tabular}  & \begin{tabular}[c]{@{}c@{}}$\textbf{dog}$\\$\textbf{[1.0000]}$\end{tabular} & \begin{tabular}[c]{@{}c@{}}$\textbf{dog}$\\$\textbf{[0.5663]}$\end{tabular} \\ \hline
		\textbf{$\text{V}_3$} & \begin{tabular}[c]{@{}c@{}}$\textbf{4}$\\$\textbf{[1.0000]}$\end{tabular} & \begin{tabular}[c]{@{}c@{}}{9}\\{[0.9994]}\end{tabular} & \begin{tabular}[c]{@{}c@{}}{9}\\{[1.0000]}\end{tabular} & \begin{tabular}[c]{@{}c@{}}$\textbf{4}$\\$\textbf{[0.9999]}$\end{tabular} & \begin{tabular}[c]{@{}c@{}}$\textbf{4}$\\$\textbf{[1.0000]}$\end{tabular} & \begin{tabular}[c]{@{}c@{}}$\textbf{4}$\\$\textbf{[1.0000]}$\end{tabular} & \begin{tabular}[c]{@{}c@{}}{9}\\{[0.9525]}\end{tabular} & \begin{tabular}[c]{@{}c@{}}$\textbf{dog}$\\$\textbf{[0.9999]}$\end{tabular} & \begin{tabular}[c]{@{}c@{}}$\text{horse}$\\$\text{[0.8076]}$\end{tabular} & \begin{tabular}[c]{@{}c@{}}$\text{horse}$\\$\text{[0.5280]}$\end{tabular} & \begin{tabular}[c]{@{}c@{}}$\textbf{dog}$\\$\textbf{[0.9934]}$\end{tabular} & \begin{tabular}[c]{@{}c@{}}$\textbf{dog}$\\$\textbf{[0.9996]}$\end{tabular} & \begin{tabular}[c]{@{}c@{}}$\textbf{dog}$\\$\textbf{[0.9996]}$\end{tabular} & \begin{tabular}[c]{@{}c@{}}$\text{cat}$\\$\text{[1.0000]}$\end{tabular} \\ \hline
		\textbf{$\text{V}_4$} & \begin{tabular}[c]{@{}c@{}}$\textbf{4}$\\$\textbf{[1.0000]}$\end{tabular} & \begin{tabular}[c]{@{}c@{}}$\textbf{4}$\\$\textbf{[0.9686]}$\end{tabular} & \begin{tabular}[c]{@{}c@{}}{9}\\{[0.8475]}\end{tabular} & \begin{tabular}[c]{@{}c@{}}$\textbf{4}$\\$\textbf{[0.8105]}$\end{tabular} & \begin{tabular}[c]{@{}c@{}}$\textbf{4}$\\$\textbf{[0.9985]}$\end{tabular} & \begin{tabular}[c]{@{}c@{}}$\textbf{4}$\\$\textbf{[0.9886]}$\end{tabular} & \begin{tabular}[c]{@{}c@{}}$\textbf{4}$\\$\textbf{[0.9640]}$\end{tabular} & \begin{tabular}[c]{@{}c@{}}$\textbf{dog}$\\$\textbf{[1.0000]}$\end{tabular} & \begin{tabular}[c]{@{}c@{}}$\textbf{dog}$\\$\textbf{[0.7023]}$\end{tabular} & \begin{tabular}[c]{@{}c@{}}$\textbf{dog}$\\$\textbf{[0.8187]}$\end{tabular} & \begin{tabular}[c]{@{}c@{}}$\textbf{dog}$\\$\textbf{[0.9998]}$\end{tabular} & \begin{tabular}[c]{@{}c@{}}$\textbf{dog}$\\$\textbf{[1.0000]}$\end{tabular} & \begin{tabular}[c]{@{}c@{}}$\textbf{dog}$\\$\textbf{[1.0000]}$\end{tabular} & \begin{tabular}[c]{@{}c@{}}$\text{cat}$\\$\text{[0.9584]}$\end{tabular} \\ \hline
		\textbf{$\text{V}_5$} & \begin{tabular}[c]{@{}c@{}}$\textbf{4}$\\$\textbf{[1.0000]}$\end{tabular} & \begin{tabular}[c]{@{}c@{}}$\text{9}$\\$\text{[0.8796]}$\end{tabular} & \begin{tabular}[c]{@{}c@{}}$\text{9}$\\$\text{[1.0000]}$\end{tabular} & \begin{tabular}[c]{@{}c@{}}$\text{9}$\\$\text{[0.5978]}$\end{tabular} & \begin{tabular}[c]{@{}c@{}}$\textbf{4}$\\$\textbf{[0.9974]}$\end{tabular} & \begin{tabular}[c]{@{}c@{}}$\textbf{4}$\\$\textbf{[0.9939]}$\end{tabular} & \begin{tabular}[c]{@{}c@{}}$\text{9}$\\$\text{[0.9130]}$\end{tabular} & \begin{tabular}[c]{@{}c@{}}$\textbf{dog}$\\$\textbf{[0.9956]}$\end{tabular}  & \begin{tabular}[c]{@{}c@{}}$\textbf{dog}$\\$\textbf{[0.9484]}$\end{tabular} & \begin{tabular}[c]{@{}c@{}}$\textbf{dog}$\\$\textbf{[0.9881]}$\end{tabular} & \begin{tabular}[c]{@{}c@{}}$\textbf{dog}$\\$\textbf{[0.9961]}$\end{tabular} & \begin{tabular}[c]{@{}c@{}}$\textbf{dog}$\\$\textbf{[0.9959]}$\end{tabular} & \begin{tabular}[c]{@{}c@{}}$\textbf{dog}$\\$\textbf{[0.7752]}$\end{tabular} & \begin{tabular}[c]{@{}c@{}}$\text{cat}$\\$\text{[0.7850]}$\end{tabular} \\ \hline
		\textbf{$\text{V}_6$} & \begin{tabular}[c]{@{}c@{}}$\textbf{4}$\\$\textbf{[0.9999]}$\end{tabular} & \begin{tabular}[c]{@{}c@{}}$\textbf{4}$\\$\textbf{[0.8277]}$\end{tabular} & \begin{tabular}[c]{@{}c@{}}$\text{8}$\\$\text{[0.9915]}$\end{tabular} & \begin{tabular}[c]{@{}c@{}}$\textbf{4}$\\$\textbf{[0.9949]}$\end{tabular} & \begin{tabular}[c]{@{}c@{}}$\textbf{4}$\\$\textbf{[0.9995]}$\end{tabular} & \begin{tabular}[c]{@{}c@{}}$\textbf{4}$\\$\textbf{[0.9939]}$\end{tabular} & \begin{tabular}[c]{@{}c@{}}$\textbf{4}$\\$\textbf{[0.9994]}$\end{tabular} & \begin{tabular}[c]{@{}c@{}}$\textbf{dog}$\\$\textbf{[1.0000]}$\end{tabular} & \begin{tabular}[c]{@{}c@{}}$\textbf{dog}$\\$\textbf{[1.0000]}$\end{tabular} & \begin{tabular}[c]{@{}c@{}}$\textbf{dog}$\\$\textbf{[1.0000]}$\end{tabular} & \begin{tabular}[c]{@{}c@{}}$\textbf{dog}$\\$\textbf{[1.0000]}$\end{tabular} & \begin{tabular}[c]{@{}c@{}}$\textbf{dog}$\\$\textbf{[1.0000]}$\end{tabular} & \begin{tabular}[c]{@{}c@{}}$\textbf{dog}$\\$\textbf{[1.0000]}$\end{tabular} & \begin{tabular}[c]{@{}c@{}}$\textbf{dog}$\\$\textbf{[0.9233]}$\end{tabular} \\ \hline
		\textbf{$\text{V}_7$} & \begin{tabular}[c]{@{}c@{}}$\textbf{4}$\\$\textbf{[1.0000]}$\end{tabular} & \begin{tabular}[c]{@{}c@{}}$\text{9}$\\$\text{[0.6400]}$\end{tabular} & \begin{tabular}[c]{@{}c@{}}$\text{9}$\\$\textbf{[0.7800]}$\end{tabular} & \begin{tabular}[c]{@{}c@{}}$\textbf{4}$\\$\textbf{[0.5576]}$\end{tabular} & \begin{tabular}[c]{@{}c@{}}$\text{3}$\\$\text{[0.9908]}$\end{tabular} & \begin{tabular}[c]{@{}c@{}}$\textbf{4}$\\$\textbf{[0.4516]}$\end{tabular} & \begin{tabular}[c]{@{}c@{}}$\textbf{4}$\\$\textbf{[0.9903]}$\end{tabular} &
		\begin{tabular}[c]{@{}c@{}}$\textbf{dog}$\\$\textbf{[1.0000]}$\end{tabular} & \begin{tabular}[c]{@{}c@{}}$\text{horse}$\\$\text{[0.9994]}$\end{tabular} & \begin{tabular}[c]{@{}c@{}}$\textbf{dog}$\\$\textbf{[0.8934]}$\end{tabular} & \begin{tabular}[c]{@{}c@{}}$\textbf{dog}$\\$\textbf{[0.9958]}$\end{tabular} & \begin{tabular}[c]{@{}c@{}}$\textbf{dog}$\\$\textbf{[0.9998]}$\end{tabular} & \begin{tabular}[c]{@{}c@{}}$\text{horse}$\\$\text{[0.8080]}$\end{tabular} & \begin{tabular}[c]{@{}c@{}}$\text{frog}$\\$\text{[0.9975]}$\end{tabular} \\ \hline
		\textbf{$\text{V}_8$} & \begin{tabular}[c]{@{}c@{}}$\textbf{4}$\\$\textbf{[0.9997]}$\end{tabular} &\begin{tabular}[c]{@{}c@{}}9\\$\text{[0.4434]}$\end{tabular}   & \begin{tabular}[c]{@{}c@{}}9\\$\text{[0.5201]}$\end{tabular} & \begin{tabular}[c]{@{}c@{}}8\\$\text{[0.8267]}$\end{tabular} & \begin{tabular}[c]{@{}c@{}}\textbf{4}\\$\textbf{[0.8710]}$\end{tabular} & \begin{tabular}[c]{@{}c@{}}3\\$\text{[0.7317]}$\end{tabular} & \begin{tabular}[c]{@{}c@{}}\textbf{4}\\$\textbf{[0.9998]}$\end{tabular} & \begin{tabular}[c]{@{}c@{}}$\textbf{dog}$\\$\textbf{[0.8284]}$\end{tabular} & \begin{tabular}[c]{@{}c@{}}$\textbf{dog}$\\$\textbf{[0.6055]}$\end{tabular} &\begin{tabular}[c]{@{}c@{}}$\textbf{dog}$\\$\textbf{[0.7699]}$\end{tabular}  & \begin{tabular}[c]{@{}c@{}}$\textbf{dog}$\\$\textbf{[0.8572]}$\end{tabular} & \begin{tabular}[c]{@{}c@{}}$\textbf{dog}$\\$\textbf{[0.8565]}$\end{tabular} & \begin{tabular}[c]{@{}c@{}}$\textbf{dog}$\\$\textbf{[0.6601]}$\end{tabular} & \begin{tabular}[c]{@{}c@{}}$\textbf{dog}$\\$\textbf{[0.5119]}$\end{tabular} \\ \hline
		\textbf{$\text{V}_9$} & \begin{tabular}[c]{@{}c@{}}\textbf{4}\\$\textbf{[0.9948]}$\end{tabular} &\begin{tabular}[c]{@{}c@{}}9\\$\text{[0.5192]}$\end{tabular} & \begin{tabular}[c]{@{}c@{}}9\\$\text{[0.6377]}$\end{tabular} & \begin{tabular}[c]{@{}c@{}}\textbf{4}\\$\textbf{[0.8961]}$\end{tabular} & \begin{tabular}[c]{@{}c@{}}\textbf{4}\\$\textbf{[0.9885]}$\end{tabular} & \begin{tabular}[c]{@{}c@{}}\textbf{4}\\$\textbf{[0.9612]}$\end{tabular} & \begin{tabular}[c]{@{}c@{}}\textbf{4}\\$\textbf{[0.3684]}$\end{tabular} & \begin{tabular}[c]{@{}c@{}}$\textbf{dog}$\\$\textbf{[0.9996]}$\end{tabular} & \begin{tabular}[c]{@{}c@{}}$\textbf{dog}$\\$\textbf{[0.9987]}$\end{tabular} & \begin{tabular}[c]{@{}c@{}}$\textbf{dog}$\\$\textbf{[0.9995]}$\end{tabular} & \begin{tabular}[c]{@{}c@{}}$\textbf{dog}$\\$\textbf{[0.9996]}$\end{tabular} & \begin{tabular}[c]{@{}c@{}}$\textbf{dog}$\\$\textbf{[0.9994]}$\end{tabular} & \begin{tabular}[c]{@{}c@{}}$\textbf{dog}$\\$\textbf{[0.9975]}$\end{tabular} & \begin{tabular}[c]{@{}c@{}}$\text{cat}$\\$\text{[0.9879]}$\end{tabular} \\ \hline
		\textbf{$\text{V}_{10}$} & \begin{tabular}[c]{@{}c@{}}\textbf{4}\\$\textbf{[0.9997]}$\end{tabular} & \begin{tabular}[c]{@{}c@{}}\text{9}\\$\text{[0.9166]}$\end{tabular} & \begin{tabular}[c]{@{}c@{}}\text{9}\\$\text{[0.9994]}$\end{tabular}  & \begin{tabular}[c]{@{}c@{}}\textbf{4}\\$\textbf{[0.3515]}$\end{tabular} & \begin{tabular}[c]{@{}c@{}}\textbf{4}\\$\textbf{[0.5573]}$\end{tabular} & \begin{tabular}[c]{@{}c@{}}\text{2}\\$\text{[0.4868]}$\end{tabular}  & \begin{tabular}[c]{@{}c@{}}\text{9}\\$\text{[0.6144]}$\end{tabular}  & \begin{tabular}[c]{@{}c@{}}$\textbf{dog}$\\$\textbf{[0.9990]}$\end{tabular} & \begin{tabular}[c]{@{}c@{}}$\textbf{dog}$\\$\textbf{[0.9985]}$\end{tabular} & \begin{tabular}[c]{@{}c@{}}$\textbf{dog}$\\$\textbf{[0.9990]}$\end{tabular} & \begin{tabular}[c]{@{}c@{}}$\textbf{dog}$\\$\textbf{[0.9989]}$\end{tabular} & \begin{tabular}[c]{@{}c@{}}$\textbf{dog}$\\$\textbf{[0.9988]}$\end{tabular} & \begin{tabular}[c]{@{}c@{}}$\textbf{dog}$\\$\textbf{[0.9989]}$\end{tabular} & \begin{tabular}[c]{@{}c@{}}$\textbf{dog}$\\$\textbf{[0.9869]}$\end{tabular} \\ \hhline{|=|=|=|=|=|=|=|=|=|=|=|=|=|=|=|}
		\textbf{\begin{tabular}[r]{@{}r@{}}Model Verification\\ Ensemble\end{tabular}} & \begin{tabular}[c]{@{}c@{}}\textbf{4}\\$\textbf{[0.9982]}$\end{tabular} & \begin{tabular}[c]{@{}c@{}}\text{9}\\$\text{[0.4663]}$\end{tabular}  & \begin{tabular}[c]{@{}c@{}}\text{9}\\$\text{[0.5459]}$\end{tabular}  & \begin{tabular}[c]{@{}c@{}}\textbf{4}\\$\textbf{[0.6303]}$\end{tabular} & \begin{tabular}[c]{@{}c@{}}\textbf{4}\\$\textbf{[0.9952]}$\end{tabular} & \begin{tabular}[c]{@{}c@{}}\textbf{4}\\$\textbf{[0.9830]}$\end{tabular} & \begin{tabular}[c]{@{}c@{}}\textbf{4}\\$\textbf{[0.4559]}$\end{tabular} & \begin{tabular}[c]{@{}c@{}}$\textbf{dog}$\\$\textbf{[0.9775]}$\end{tabular} & \begin{tabular}[c]{@{}c@{}}$\textbf{dog}$\\$\textbf{[0.6748]}$\end{tabular} & \begin{tabular}[c]{@{}c@{}}$\textbf{dog}$\\$\textbf{[0.8246]}$\end{tabular} & \begin{tabular}[c]{@{}c@{}}$\textbf{dog}$\\$\textbf{[0.9722]}$\end{tabular} & \begin{tabular}[c]{@{}c@{}}$\textbf{dog}$\\$\textbf{[0.9599]}$\end{tabular} & \begin{tabular}[c]{@{}c@{}}$\textbf{dog}$\\$\textbf{[0.8334]}$\end{tabular} & \begin{tabular}[c]{@{}c@{}}$\text{cat}$\\$\text{[0.3731]}$\end{tabular} \\ \hline
		\textbf{\begin{tabular}[r]{@{}r@{}}Denoising-Verification\\ Cross-Layer Ensemble\end{tabular}} & \begin{tabular}[c]{@{}c@{}}\textbf{4}\\$\textbf{[1.0000]}$\end{tabular} & \begin{tabular}[c]{@{}c@{}}\textbf{4}\\$\textbf{[0.9825]}$\end{tabular} &\begin{tabular}[c]{@{}c@{}}\textbf{4}\\$\textbf{[0.8452]}$\end{tabular}  & \begin{tabular}[c]{@{}c@{}}\textbf{4}\\$\textbf{[0.9994]}$\end{tabular} & \begin{tabular}[c]{@{}c@{}}\textbf{4}\\$\textbf{[0.9996]}$\end{tabular} & \begin{tabular}[c]{@{}c@{}}\textbf{4}\\$\textbf{[0.9996]}$\end{tabular} & \begin{tabular}[c]{@{}c@{}}\textbf{4}\\$\textbf{[0.9586]}$\end{tabular} & \begin{tabular}[c]{@{}c@{}}$\textbf{dog}$\\$\textbf{[0.9844]}$\end{tabular} & \begin{tabular}[c]{@{}c@{}}$\textbf{dog}$\\$\textbf{[0.9576]}$\end{tabular} & \begin{tabular}[c]{@{}c@{}}$\textbf{dog}$\\$\textbf{[0.9775]}$\end{tabular} & \begin{tabular}[c]{@{}c@{}}$\textbf{dog}$\\$\textbf{[0.9809]}$\end{tabular} & \begin{tabular}[c]{@{}c@{}}$\textbf{dog}$\\$\textbf{[0.9798]}$\end{tabular} & \begin{tabular}[c]{@{}c@{}}$\textbf{dog}$\\$\textbf{[0.9742]}$\end{tabular} & \begin{tabular}[c]{@{}c@{}}$\textbf{dog}$\\$\textbf{[0.9306]}$\end{tabular} \\ \hline
	\end{tabular}
\vspace{-4mm}
\end{table*}

Given a query example $\boldsymbol{x}$, a model verification ensemble of size $Z_\mathcal{V}$ is selected randomly at runtime from the $\kappa$-ranked list of ensembles as the team of verifiers for the target model, each member, denoted by $\text{V}_i$, takes the query example as the input and produces a prediction probability distribution, $\text{V}_i(\boldsymbol{x}):\mathbb{R}^D\rightarrow\mathbb{R}^K$, over the output space. Then, the \scheme{}-approved prediction output $\hat{y}$ will be produced based on a soft-voting scheme. In addition to the unweighted majority voting (recall the denoising ensemble in Section~\ref{sec:dnn-denoising}), another example of such voting scheme could be weighted averaging based on the confidence of the prediction made by each verifier in the chosen verification ensemble: 
\vspace{-1mm}
\begin{equation}\small
\hat{y}=\arg\max{}_{1\leq j\leq K}\frac{1}{Z_\mathcal{V}}\sum_{i=1}^{Z_\mathcal{V}}\text{V}_{i,j}(\boldsymbol{x}),
\end{equation}
where $\text{V}_{i,j}(\boldsymbol{x})$ denotes the confidence of $\boldsymbol{x}$ being from class $j$ predicted by verifier $\text{V}_i$. The 2nd row from the bottom of Table~\ref{fig:verification-example} shows the results of using model verification ensemble defense with the $3$-model team $\text{V}_5$,$\text{V}_6$,$\text{V}_9$ for MNIST and the $10$-model team $\text{V}_1, ..., \text{V}_{10}$ for CIFAR-10. Without using the model denoising ensemble, we observe that for these two query inputs (digit $4$ and dog), most of the adversarial examples can be correctly classified, including JSMA and CW$_{\infty}$ attacks on the MNIST example, the FGSM and BIM attacks on the CIFAR-10 example, which failed by both the denoising ensemble and any of the two denoisers in Figure~\ref{fig:denoising-example}. However, the verification ensemble selected from the ten base models still fails to repair the adversarial examples generated by FGSM and BIM for digit $4$ of MNIST and adversarial example generated by JSMA for the dog example of CIFAR-10. We are motivated to combine denoising ensemble with verification ensemble, which has a high probability to outperform denoising ensemble or verification ensemble alone. 

\vspace{-2mm}
\subsection{Denoising-Verification Cross-Layer Ensemble}\label{sec:cross-layer}
There are a number of ways that we can create cross-layer ensembles. Due to the space constraint, we below describe two approaches representing two ends of the spectrum. The first approach sends an ensemble voted denoising output to the model verification ensemble. The second approach sends to the model verification ensemble all denoised versions of the query input produced by every member of the denoising ensemble. Let the back-end defense to conduct the cross-layer ensemble. This approach ensures that no errors from the front-end defense will be propagated to the back-end defense phase, at the cost of sending every denoised version. Other solutions could be in between of these two spectrums, such as sending only those voted denoising outputs that are ranked higher than the system-defined consensus threshold.

{\bf One-to-Many Denoising-Verification Cross-Layer Ensemble.\/}
In the front-end defense phase, \scheme{} will employ denoising ensemble to produce one transformed input example by removing adversarial noises through unsupervised denoising and multiple model denoising consensus, such as unweighted majority voting, simple averaging or weighted averaging. In the back-end defense phase, it takes the voted denoised version of the query example, performs the target model prediction first and then performs the model verification ensemble over the target prediction outcome. This process serves two purposes: (1) It aims to ensure the correctness of the target model prediction outcome by repairing the prediction error through the cross-layer ensemble. (2) It also provides the detection capability to flag those adversarial examples that escape from or cannot be repaired by the cross-layer model ensemble defense framework. 

{\bf Many-to-Many Denoising-Verification Cross-Layer Ensemble.\/}
When the denoising ensemble as the front-end defense chooses not to filter out any denoised versions of the query input. The back-end defense phase will need to consider each denoised version. In this case, we may choose to produce one cross-layer verified prediction outcome for each denoised version and then use an ensemble ranking algorithm, such as unweighted majority voting, simple averaging or weighted averaging, to determine the \scheme{} defense verified prediction outcome for the target model. Alternatively, we could also pool all the cross-layer verification results for all denoised versions together and run the ensemble consensus algorithm to rank and select the top-$1$ or top-$k$ prediction outcomes. 

\begin{table*}[b]
	\scriptsize
		\renewcommand\thetable{V}
	\centering
	\caption{Comparing the \scheme{} cross-layer ensemble defense with three representative defenses on benign accuracy and DSR.}
	\vspace{-2mm}
	\label{tab:exp-comparison}
	\begin{tabular}{|c|l|c|c|c|c|c|c|c|c|c|c|c|c|c|}
		\hline
		\multirow{2}{*}{} & \multicolumn{1}{c|}{\multirow{2}{*}{\textbf{Defense}}} & \multirow{2}{*}{\textbf{\begin{tabular}[c]{@{}c@{}}Benign\\ Accuracy\end{tabular}}} & \textbf{FGSM} & \textbf{BIM} & \multicolumn{2}{c|}{\textbf{CW$_\infty$}} & \multicolumn{2}{c|}{\textbf{CW$_2$}} & \multicolumn{2}{c|}{\textbf{CW$_0$}} & \multicolumn{2}{c|}{\textbf{JSMA}} & \multirow{2}{*}{\textbf{Average}} & \multirow{2}{*}{\textbf{Std}} \\ \cline{4-13}
		& \multicolumn{1}{c|}{} &  & \multicolumn{2}{c|}{\textbf{UA}} & \textbf{ML} & \textbf{LL}  & \textbf{ML} & \textbf{LL} & \textbf{ML} & \textbf{LL} & \textbf{ML} & \textbf{LL} &  & \\ \hline
		\multirow{4}{*}{\rot{MNIST}} & MODEF & $0.9855$ & $0.89$ & $\mathbf{0.86}$ & $\mathbf{0.99}$ & $\mathbf{0.97}$ & $\mathbf{0.98}$ & $\mathbf{0.95}$ & $\mathbf{0.94}$ & $\mathbf{0.92}$ & $\mathbf{0.98}$ & $\mathbf{0.90}$ & $\mathbf{0.94}$ & $\mathbf{0.04}$\\ \cline{2-15} 
		& Adversarial Training & $\mathbf{0.9884}$ & $\mathbf{0.91}$ & $0.81$ & $0.97$ & $0.88$ & $0.92$ & $0.84$ & $0.67$ & $0.64$ & $0.73$ & $0.69$ & $0.81$ & $0.12$ \\ \cline{2-15} 
		& Defensive Distillation & $0.9784$ & $0.68$ & $0.57$ & $0.91$ & $0.85$ & $0.91$ & $0.84$ & $0.78$ & $0.72$ & $0.85$ & $0.75$ & $0.79$ & $0.11$ \\ \cline{2-15} 
		& Ensemble Transformation & $0.9820$ & $0.60$ & $0.22$ & $0.64$ & $0.51$ & $0.37$ & $0.33$ & $0.21$ & $0.21$ & $0.57$ & $0.64$ & $0.43$ & $0.18$ \\ \hhline{|=|=|=|=|=|=|=|=|=|=|=|=|=|=|=|}
		\multirow{4}{*}{\rot{CIFAR-10}} & MODEF & $\mathbf{0.9631}$ & $\mathbf{0.91}$ & $\mathbf{0.96}$ & $\mathbf{0.98}$ & $\mathbf{0.98}$ & $\mathbf{0.98}$ & $\mathbf{0.98}$ & $\mathbf{0.94}$ & $\mathbf{0.93}$ & $\mathbf{0.92}$ & $\mathbf{0.80}$ & $\mathbf{0.94}$ & $\mathbf{0.06}$ \\ \cline{2-15} 
		& Adversarial Training & $0.8790$ & $0.64$ & $0.58$ & $0.68$ & $0.77$  & $0.75$ & $0.79$ & $0.44$ & $0.48$ & $0.50$ & $0.45$ & $0.61$ & $0.14$ \\ \cline{2-15} 
		& Defensive Distillation & $0.9118$ & $0.60$ & $0.65$ & $0.79$ & $0.88$ & $0.86$ & $0.90$ & $0.60$ & $0.69$ & $0.70$ & $0.47$ & $0.71$ & $0.14$ \\ \cline{2-15} 
		& Ensemble Transformation & $0.8014$ & $0.23$ & $0.40$ & $0.56$ & $0.61$ & $0.57$ & $0.61$ & $0.19$ & $0.34$ & $0.45$ & $0.41$ & $0.44$ & $0.15$ \\ \hline
	\end{tabular}
\end{table*}
Concretely, given a query example $\boldsymbol{x}$ at runtime, we first exercise the front-end defense by employing a model denoising ensemble of size $Z_\mathcal{D}$ to produce $Z_\mathcal{D}$ denoised versions of $\boldsymbol{x}$, denoted by $\mathcal{D}_1(\boldsymbol{x}),...\mathcal{D}_{Z_D}(\boldsymbol{x})$. Then, we select a model verification ensemble team of size $Z_\mathcal{V}$ from the $\kappa$-ranked list of model verification ensembles. For each of the denoised versions, we will produce $Z_\mathcal{D}$ probability distributions $P_1(\boldsymbol{x}),...,P_{Z_{\mathcal{D}}}(\boldsymbol{x})$ where $P_k(\boldsymbol{x})=\frac{1}{Z_\mathcal{V}}\sum_{i=1}^{Z_\mathcal{V}}\text{V}_i(\mathcal{D}_k(\boldsymbol{x}))$. We select the one, denoted as the $c$-th denoiser, producing the most confident prediction and take the corresponding predicted class label as the final defense-approved prediction. Similar to the one-to-many approach, detection capability can be offered to flag irreparable adversarial examples. 

Table~\ref{fig:verification-example} shows the result of the cross-layer ensemble approach in the last row using the above many-to-many cross-layer ensemble defense. Compared with model verification ensemble only defense (the 2nd row from the bottom), clearly, the denoising-verification cross-layer ensemble can successfully verify and repair all the adversarial examples. In this experiment, the cross-layer ensemble defense method uses the two denoisers in Figure~\ref{fig:denoising-example}. 

\vspace{-1mm}
\section{Experimental Evaluation}\label{sec:experimental-results}
\vspace{-1mm}
\subsection{Experimental Setup}
We evaluate \scheme{} on two popular benchmark datasets: MNIST and CIFAR-10. Each of them is associated with one denoiser trained with Gaussian noise and another trained with salt-and-pepper noise. Their neural architectures are reported in Table~\ref{tab:a-denoiser-architectures} where we use {[}kernel width{]}$\times${[}output channel{]} to denote parameters of convolution layers and {[}kernel width{]} to denote parameters of average pooling and upsampling layers.
The volume of Gaussian noise is set to be $0.3$ for MNIST and $0.01$ for CIFAR-10 while the ratio of pixels randomly modified by the salt-and-pepper corruption process is set to be $0.1$ for both datasets. All denoisers are trained with Adam optimizer with a learning rate of $0.001$ and a batch size of $256$. Euclidean distance is employed to measure the difference between two examples (i.e., the function $d$ in Equation~\ref{eq:dae-opt}) 
while the regularization hyperparameter $\lambda$ is set to be $10^{-9}$. 
\begin{table}[h]
    \scriptsize
    \centering
	\renewcommand\thetable{IV}
	\caption{The neural network structures of DNN denoisers adopted in MNIST and CIFAR-10 experiments.}
	\vspace{-2mm}
	\label{tab:a-denoiser-architectures}
	\begin{tabular}{|l|l|lll}
		\cline{1-2} \cline{4-5}
		\multicolumn{2}{|c|}{\textbf{MNIST}} & \multicolumn{1}{c|}{\textbf{}} & \multicolumn{2}{c|}{\textbf{CIFAR-10}} \\ \cline{1-2} \cline{4-5} 
		\multicolumn{1}{|c|}{\textbf{Layer}} & \multicolumn{1}{c|}{\textbf{Settings}} & \multicolumn{1}{c|}{\textbf{}} & \multicolumn{1}{c|}{\textbf{Layer}} & \multicolumn{1}{c|}{\textbf{Settings}} \\ \cline{1-2} \cline{4-5} 
		Convolution + Sigmoid & $3\times3\times3$ & \multicolumn{1}{l|}{} & \multicolumn{1}{l|}{Convolution + Sigmoid} & \multicolumn{1}{l|}{$3\times3\times3$} \\ \cline{1-2} \cline{4-5} 
		Average Pooling & $2\times2$ & \multicolumn{1}{l|}{} & \multicolumn{1}{l|}{Convolution + Sigmoid} & \multicolumn{1}{l|}{$3\times3\times3$} \\ \cline{1-2} \cline{4-5} 
		Convolution + Sigmoid & $3\times3\times3$ & \multicolumn{1}{l|}{} & \multicolumn{1}{l|}{Convolution + Sigmoid} & \multicolumn{1}{l|}{$3\times3\times1$} \\ \cline{1-2} \cline{4-5} 
		Convolution + Sigmoid & $3\times3\times3$ &  &  &  \\ \cline{1-2}
		Upsampling & $2\times2$ &  &  &  \\ \cline{1-2}
		Convolution + Sigmoid & $3\times3\times3$ &  &  &  \\ \cline{1-2}
		Convolution + Sigmoid & $3\times3\times1$ &  &  &  \\ \cline{1-2}
	\end{tabular}
	\vspace{-2mm}
\end{table} 
The number of training epochs for MNIST is $100$ and that for CIFAR-10 is $400$. 
For each dataset, a set of ten model verifiers is collected from the public domain. As reported in Table~\ref{tab:a-verifier-acc}, each of them provides a competitive accuracy on benign examples as the target model. In our experimental studies, we consider a fixed team of DNN denoisers and demonstrate different teaming options through model verifiers. All experiments were run on Google Colab operated under Ubuntu 18.04.2 LTS with an Intel Xeon CPU (two cores @2.3 GHz), an NVIDIA Tesla K80 (12 GB) GPU, and 13 GB RAM.

\vspace{-1mm}
\subsection{Comparing \scheme{} with Existing Representative Defenses}
This set of experiments compares the \scheme{} cross-layer ensemble defense with three representative defense methods: adversarial training~\cite{szegedy2013intriguing}, defensive distillation~\cite{papernot2016distillation}, and ensemble transformation of input examples~\cite{guo2017countering}. We select the ensemble team to be $\text{V}_5$,$\text{V}_6$,$\text{V}_9$ for MNIST and $\text{V}_1,...,\text{V}_{10}$ for CIFAR-10 from the $\kappa$-ranked list based on their model diversity. They are exploited by default in the following studies unless further specified. Table~\ref{tab:exp-comparison} reports the results in benign accuracy under no attack and DSR under ten attacks for MNIST and CIFAR-10 introduced in Section~\ref{sec:attacks}.
Clearly, \scheme{} outperforms existing schemes for all attacks on CIFAR-10 and nine out of ten attacks on MNIST. The one exception on MNIST is the FGSM attack, under which \scheme{} has a DSR of $0.89$, slightly lower than the DSR of $0.91$ by adversarial training. However, adversarial training is attack-dependent and does not generalize over attack algorithms. Indeed, existing defense methods tend to perform inconsistently across different attacks. For instance, defensive distillation reaches a high DSR of $0.90$ under the CW$_2$ LL attack on CIFAR-10 but achieves only a low DSR of $0.47$ under JSMA LL. This can also be observed in their high standard deviation of DSRs over attacks. In contrast, \scheme{} achieves a competitive benign accuracy with a high and stable DSR, showing that it can generalize well across all attacks. 
\vspace{-1mm}
\subsection{Comparison of Three MODEF Ensemble Defenses}
Given that \scheme{} outperforms existing defenses, the next set of experiments was conducted to compare the performance of three \scheme{} ensemble approaches and understand how denoising ensemble and verification ensemble complement one another in delivering higher robustness for the target model in the cross-layer defense. Table~\ref{tab:exp-denoiser} reports the results with upper wide table for MNIST and lower wide table for CIFAR-10. 

{\bf Denoising Ensemble Defense.\/}
We have shown in Section~\ref{sec:attacks} that most of the attacks successfully fool the target model by $100\%$ attack success rate or by reducing its test accuracy significantly to zero. Table~\ref{tab:exp-denoiser} shows three important results: First, by employing a DNN denoiser, either trained with Gaussian or salt-and-pepper noise, one can improve the robustness of the target model, as shown in the columns under ``Denoiser (Gaussian Noise)'' and ``Denoiser (Salt-and-Pepper Noise)''. Second, even though some attacks are mitigated with a high DSR, the effectiveness of one denoiser varies wildly across different attacks and datasets. For instance, the Gaussian denoiser on MNIST achieves a high DSR of $0.99$ in defending the CW$_\infty$ LL attack but only gets a low DSR of $0.28$ under the CW$_0$ ML attack. In fact, we have performed experiments on multiple denoisers, and we found consistently that no DNN denoiser can remove adversarial perturbations from all eleven attacks examined and each denoiser fails to generalize well over all attacks. Third, the DSR of model denoising ensemble is consistently better than using one denoiser as shown in the columns under ``Model Denoising Ensemble Defense''. This validates our argument that a robust defense method should not rely on a single defense strategy, such as one denoiser as adopted in prior works~\cite{sahay2019combatting, liao2018defense} but employ a denoising ensemble. Also, a simple denoising ensemble as those advocated in~\cite{meng2017magnet} may not outperform a diversity optimized denoising ensemble, as an ensemble team of diverse denoisers offers different denoising effects and hence generalizes better over different adversarial attacks.

\begin{table*}

	\scriptsize
	\centerfloat
	\caption{Comparing defensibility of the three \scheme{} on MNIST and CIFAR-10.}
	\vspace{-2mm}
	\label{tab:exp-denoiser}
	\begin{tabular}{|c|c|c|c|c|c|c|c|c|c|c|c|c|c|c|c|c|c|c|c|c|c|}
		\hline
		\multicolumn{2}{|c|}{\textbf{}} & \multicolumn{4}{c|}{\textbf{\begin{tabular}[c]{@{}c@{}}Denoiser\\ (Gaussian Noise)\end{tabular}}} & \multicolumn{4}{c|}{\textbf{\begin{tabular}[c]{@{}c@{}}Denoiser\\ (Salt-and-Pepper Noise)\end{tabular}}} & \multicolumn{4}{c|}{\textbf{\begin{tabular}[c]{@{}c@{}}Model Denoising\\ Ensemble Defense\end{tabular}}} & \multicolumn{4}{c|}{\textbf{\begin{tabular}[c]{@{}c@{}}Model Verification\\ Ensemble Defense\end{tabular}}} & \multicolumn{4}{c|}{\textbf{\begin{tabular}[c]{@{}c@{}}Denoising-Verification Cross-Layer\\Ensemble Defense\end{tabular}}} \\ \hline
		\multicolumn{2}{|c|}{\textbf{MNIST}} & \textbf{PSR} & \textbf{TSR} & \textbf{DSR} & \textbf{FP} & \textbf{PSR} & \textbf{TSR} & \textbf{DSR} & \textbf{FP} & \textbf{PSR} & \textbf{TSR} & \textbf{DSR} & \textbf{FP} & \textbf{PSR} & \textbf{TSR} & \textbf{DSR} & \textbf{FP} & \textbf{PSR} & \textbf{TSR} & \textbf{DSR} & \textbf{FP} \\ \hline
		\multicolumn{2}{|c|}{Benign} & $\mathbf{0.9943}$ & $0.00$ & ${0.9943}$ & $0.00$ & $\mathbf{0.9943}$ & $0.00$ & ${0.9943}$ & $0.00$ & $0.9911$ & $0.0038$ & $0.9949$ & $0.0024$ & $0.9888$ & $0.00$ & $0.9888$ & $0.00$ & $0.9855$ & $0.0035$ & $0.9890$ & $0.0030$ \\ \hline
		FGSM & \multirow{2}{*}{UA} & $0.82$ & $0.00$ & $0.82$ & $0.00$ & $0.40$ & $0.00$ & $0.40$ & $0.00$ & $0.40$ & $0.43$ & $0.83$ & $0.28$ & $0.71$ & $0.00$ & $0.71$ & $0.00$ & $0.87$ & $0.02$ & $\mathbf{0.89}$ & $0.00$ \\ \cline{1-1} \cline{3-22} 
		BIM &  & $0.71$ & $0.00$ & $0.71$ & $0.00$ & $0.06$ & $0.00$ & $0.06$ & $0.00$ & $0.06$ & $0.65$ & $0.71$ & $0.17$ & $0.78$ & $0.00$ & $0.78$ & $0.00$ & $0.84$ & $0.02$ & $\mathbf{0.86}$ & $0.00$ \\ \hline
		\multirow{2}{*}{CW$_\infty$} & ML & $0.94$ & $0.00$ & $0.94$ & $0.00$ & $0.03$ & $0.00$ & $0.03$ & $0.00$ & $0.02$ & $0.93$ & $0.95$ & $0.55$ & $0.93$ & $0.00$ & $0.93$ & $0.00$ & $0.99$ & $0.00$ & $\mathbf{0.99}$ & $0.00$ \\ \cline{2-22} 
		& LL & $0.99$ & $0.00$ & $\mathbf{0.99}$ & $0.00$ & $0.02$ & $0.00$ & $0.02$ & $0.00$ & $0.02$ & $0.97$ & $\mathbf{0.99}$ & $0.62$ & $0.85$ & $0.00$ & $0.85$ & $0.00$ & $0.97$ & $0.00$ & ${0.97}$ & $0.00$ \\ \hline
		\multirow{2}{*}{CW$_2$} & ML & $0.89$ & $0.00$ & $0.89$ & $0.00$ & $0.14$ & $0.00$ & $0.14$ & $0.00$ & $0.10$ & $0.83$ & $0.93$ & $0.44$ & $0.96$ & $0.00$ & $0.96$ & $0.00$ & $0.98$ & $0.00$ & $\mathbf{0.98}$ & $0.00$ \\ \cline{2-22} 
		& LL & $0.83$ & $0.00$ & $0.83$ & $0.00$ & $0.12$ & $0.00$ & $0.12$ & $0.00$ & $0.12$ & $0.83$ & $\mathbf{0.95}$ & $0.50$ & $0.94$ & $0.00$ & $0.94$ & $0.00$ & $0.95$ & $0.00$ & $\mathbf{0.95}$ & $0.00$ \\ \hline
		\multirow{2}{*}{CW$_0$} & ML & $0.28$ & $0.00$ & $0.28$ & $0.00$ & $0.63$ & $0.00$ & $0.63$ & $0.00$ & $0.26$ & $0.40$ & $0.66$ & $0.18$ & $0.94$ & $0.00$ & $\mathbf{0.94}$ & $0.00$ & $0.94$ & $0.00$ & $\mathbf{0.94}$ & $0.00$ \\ \cline{2-22} 
		& LL & $0.41$ & $0.00$ & $0.41$ & $0.00$ & $0.68$ & $0.00$ & $0.68$ & $0.00$ & $0.41$ & $0.41$ & $0.82$ & $0.15$ & $0.89$ & $0.00$ & $0.89$ & $0.00$ & $0.92$ & $0.00$ & $\mathbf{0.92}$ & $0.00$ \\ \hline
		\multirow{2}{*}{JSMA} & ML & $0.82$ & $0.00$ & $0.82$ & $0.00$ & $0.94$ & $0.00$ & $0.94$ & $0.00$ & $0.80$ & $0.16$ & $0.96$ & $0.12$ & $0.98$ & $0.00$ & $\mathbf{0.98}$ & $0.00$ & $0.98$ & $0.00$ & $\mathbf{0.98}$ & $0.00$ \\ \cline{2-22} 
		& LL & $0.65$ & $0.00$ & $0.65$ & $0.00$ & $0.81$ & $0.00$ & $0.81$ & $0.00$ & $0.64$ & $0.22$ & $0.86$ & $0.14$ & $0.88$ & $0.00$ & $0.88$ & $0.00$ & $0.90$ & $0.00$ & $\mathbf{0.90}$ & $0.00$ \\ \hhline{|=|=|=|=|=|=|=|=|=|=|=|=|=|=|=|=|=|=|=|=|=|=|}
		\multicolumn{2}{|c|}{Average} & $0.73$ & $0.00$ & $0.73$ & $0.00$ & $0.38$ & $0.00$ & $0.38$ & $0.00$ & $0.28$ & $0.58$ & $0.86$ & $0.32$ & $0.89$ & $0.00$ & $0.89$ & $0.00$ & $0.93$ & $0.00$ & $\mathbf{0.94}$ & $0.00$ \\ \hline
		\multicolumn{2}{|c|}{Std} & $0.23$ & $0.00$ & $0.23$ & $0.00$ & $0.35$ & $0.00$ & $0.35$ & $0.00$ & $0.26$ & $0.28$ & $0.11$ & $0.18$ & $0.09$ & $0.00$ & $0.09$ & $0.00$ & $0.05$ & $0.01$ & $\mathbf{0.04}$ & $0.00$ \\ \hline
		\multicolumn{22}{|c|}{} \\ \hline
		\multicolumn{2}{|c|}{\textbf{CIFAR-10}} & \textbf{PSR} & \textbf{TSR} & \textbf{DSR} & \textbf{FP} & \textbf{PSR} & \textbf{TSR} & \textbf{DSR} & \textbf{FP} & \textbf{PSR} & \textbf{TSR} & \textbf{DSR} & \textbf{FP} & \textbf{PSR} & \textbf{TSR} & \textbf{DSR} & \textbf{FP} & \textbf{PSR} & \textbf{TSR} & \textbf{DSR} & \textbf{FP} \\ \hline
		\multicolumn{2}{|c|}{Benign} & $0.9304$ & $0.00$ & $0.9304$ & $0.00$ & $0.8835$ & $0.00$ & $0.8835$ & $0.00$ & $0.8772$ & $0.0656$ & $0.9428$ & $0.0369$ & ${0.9608}$ & $0.00$ & ${0.9608}$ & $0.00$ & $\mathbf{0.9631}$ & $0.0002$ & $0.9633$ & $0.0002$ \\ \cline{1-1} \cline{2-22} 
		FGSM & \multirow{2}{*}{UA} & $0.19$ & $0.00$ & $0.19$ & $0.00$ & $0.34$ & $0.00$ & $0.34$ & $0.00$ & $0.17$ & $0.24$ & $0.41$ & $0.12$ & $0.85$ & $0.00$ & $0.85$ & $0.00$ & $0.83$ & $0.08$ & $\mathbf{0.91}$ & $0.00$ \\ \cline{1-1} \cline{3-22} 
		BIM &  & $0.22$ & $0.00$ & $0.22$ & $0.00$ & $0.50$ & $0.00$ & $0.50$ & $0.00$ & $0.21$ & $0.33$ & $0.54$ & $0.13$ & $0.92$ & $0.00$ & $0.92$ & $0.00$ & $0.95$ & $0.01$ & $\mathbf{0.96}$ & $0.00$ \\ \hline
		\multirow{2}{*}{CW$_\infty$} & ML & $0.21$ & $0.00$ & $0.21$ & $0.00$ & $0.67$ & $0.00$ & $0.67$ & $0.00$ & $0.21$ & $0.47$ & $0.68$ & $0.22$ & $0.95$ & $0.00$ & $0.95$ & $0.00$ & $0.98$ & $0.00$ & $\mathbf{0.98}$ & $0.00$ \\ \cline{2-22} 
		& LL & $0.83$ & $0.00$ & $0.83$ & $0.00$ & $0.85$ & $0.00$ & $0.85$ & $0.00$ & $0.77$ & $0.17$ & $0.94$ & $0.11$ & $0.98$ & $0.00$ & $\mathbf{0.98}$ & $0.00$ & $0.98$ & $0.00$ & $\mathbf{0.98}$ & $0.00$ \\ \hline
		DF & UA & $0.64$ & $0.00$ & $0.64$ & $0.00$ & $0.78$ & $0.00$ & $0.78$ & $0.00$ & $0.59$ & $0.25$ & $0.84$ & $0.15$ & $0.97$ & $0.00$ & $0.97$ & $0.00$ & $0.98$ & $0.00$ & $\mathbf{0.98}$ & $0.00$ \\ \hline
		\multirow{2}{*}{CW$_2$} & ML & $0.42$ & $0.00$ & $0.42$ & $0.00$ & $0.75$ & $0.00$ & $0.75$ & $0.00$ & $0.42$ & $0.34$ & $0.76$ & $0.22$ & $0.95$ & $0.00$ & $0.95$ & $0.00$ & $0.98$ & $0.00$ & $\mathbf{0.98}$ & $0.00$ \\ \cline{2-22} 
		& LL & $0.87$ & $0.00$ & $0.87$ & $0.00$ & $0.90$ & $0.00$ & $0.90$ & $0.00$ & $0.82$ & $0.14$ & $0.96$ & $0.09$ & $0.97$ & $0.00$ & ${0.97}$ & $0.00$ & $0.98$ & $0.00$ & $\mathbf{0.98}$ & $0.00$ \\ \hline
		\multirow{2}{*}{CW$_0$} & ML & $0.03$ & $0.00$ & $0.03$ & $0.00$ & $0.33$ & $0.00$ & $0.33$ & $0.00$ & $0.03$ & $0.33$ & $0.36$ & $0.10$ & $0.88$ & $0.00$ & $0.88$ & $0.00$ & $0.86$ & $0.08$ & $\mathbf{0.94}$ & $0.00$ \\ \cline{2-22} 
		& LL & $0.21$ & $0.00$ & $0.21$ & $0.00$ & $0.70$ & $0.00$ & $0.70$ & $0.00$ & $0.18$ & $0.76$ & $\mathbf{0.94}$ & $0.34$ & $0.91$ & $0.00$ & $0.91$ & $0.00$ & $0.93$ & $0.00$ & ${0.93}$ & $0.00$ \\ \hline
		\multirow{2}{*}{JSMA} & ML & $0.40$ & $0.00$ & $0.40$ & $0.00$ & $0.77$ & $0.00$ & $0.77$ & $0.00$ & $0.39$ & $0.45$ & $0.84$ & $0.31$ & $0.83$ & $0.00$ & $0.83$ & $0.00$ & $0.92$ & $0.00$ & $\mathbf{0.92}$ & $0.00$ \\ \cline{2-22} 
		& LL & $0.32$ & $0.00$ & $0.32$ & $0.00$ & $0.71$ & $0.00$ & $0.71$ & $0.00$ & $0.31$ & $0.58$ & $\mathbf{0.89}$ & $0.38$ & $0.54$ & $0.00$ & $0.54$ & $0.00$ & $0.79$ & $0.01$ & ${0.80}$ & $0.00$ \\ \hhline{|=|=|=|=|=|=|=|=|=|=|=|=|=|=|=|=|=|=|=|=|=|=|}
		\multicolumn{2}{|c|}{Average} & $0.39$ & $0.00$ & $0.39$ & $0.00$ & $0.66$ & $0.00$ & $0.66$ & $0.00$ & $0.37$ & $0.37$ & $0.74$ & $0.20$ & $0.89$ & $0.00$ & $0.89$ & $0.00$ & $0.93$ & $0.02$ & $\mathbf{0.94}$ & $0.00$ \\ \hline
		\multicolumn{2}{|c|}{Std} & $0.27$ & $0.00$ & $0.27$ & $0.00$ & $0.19$ & $0.00$ & $0.19$ & $0.00$ & $0.25$ & $0.18$ & $0.21$ & $0.10$ & $0.12$ & $0.00$ & $0.12$ & $0.00$ & $0.07$ & $0.03$ & $\mathbf{0.05}$ & $0.00$ \\ \hline
	\end{tabular}
	\vspace{-4mm}
\end{table*}
{\bf Verification Ensemble and Cross-Layer Ensemble Defense.\/}
Table~\ref{tab:exp-denoiser} reports the defensibility of model verification ensemble under the column ``Model Verification Ensemble Defense''. Compared with denoising ensemble, although verification ensemble achieves a higher average DSR over all eleven attacks in both datasets (i.e., $0.89$ v.s. $0.86$ for MNIST and $0.89$ v.s. $0.74$ for CIFAR-10 shown on their respective ``Average'' row in Table~\ref{tab:exp-denoiser}), it performs much worse than denoising ensemble under the JSMA LL attack ($0.54$ v.s. $0.89$) for CIFAR-10 and CW$_\infty$ LL attack ($0.85$ v.s. $0.99$) for MNIST. The model verification ensemble is hence a competitive alternative to the denoising ensemble, since neither of them is the winner over all attacks on both datasets, and yet each of them is the winner for some attacks. Compared with denoising or verification ensemble only defense, the cross-layer ensemble defense consistently outperforms across almost all attacks with a remarkably small standard deviation of DSRs (see the columns of ``Denoising-Verification Cross-Layer Ensemble Defense'' in Table~\ref{tab:exp-denoiser}). In other words, the cross-layer ensemble can generalize well over the eleven attacks examined.

We make another interesting observation that although the model verification ensemble and the cross-layer ensemble may have slightly lower benign accuracy ($0.9888$ and $0.9855$) than the target model on MNIST ($0.9943$) under no attack, their benign accuracy on CIFAR-10 are $0.9608$ and $0.9631$, which are higher than the target model accuracy of $0.9484$.  This is consistent with the theoretical proof in \cite{Holloway2007-ensemble} that ensemble can reduce uncorrelated errors. For example, if we have $5$ completely independent classifiers and we use unweighted majority voting as a consensus algorithm, by the probability formula, with accuracy of $70\%$ for each classifier, we have $83.7\%$ ensemble accuracy with majority voting and if we have $101$ such classifiers, we reach $99.9\%$ ensemble accuracy by majority voting. Equivalently, let $\delta$ be the classification accuracy of each member in the ensemble $E$ of size $Z$, the accuracy can be boosted to $P(E\geq\left\lceil Z/2\right\rceil)=\sum_{i=\left\lceil Z/2\right\rceil}^{Z}{Z \choose i}\delta^i(1-\delta)^{Z-i}$. 

In \scheme{}, we consider three options to create an ensemble team of base models: (1) randomly selecting a team from the entire pool of ensemble combinations (\textbf{Rand}); (2) randomly selecting an ensemble team from the $\kappa$-ranked list (\textbf{Rand$\boldsymbol{\kappa}$}), which filters out those ensemble team combos that have high kappa values above the system-defined threshold (due to the space, the factors impact on the decision for the threshold setting is omitted here); and (3) selecting the ensemble team from the $\kappa$-ranked list with the highest prediction accuracy (\textbf{Best$\boldsymbol{\kappa}$}). The $\kappa$-ranked list is built on the kappa value of each pair of model verifiers. Different levels of disagreement occur between a pair of model verifiers. Table~\ref{fig:kappa-mnist} gives an example of the pairwise kappa values of the model verifiers on MNIST where the pair of $\text{V}_4$ and $\text{V}_5$ has a low kappa value of $0.25$, showing their high prediction disagreement diversity (high failure-independence and low error correlation). 
\begin{table}
	\centering
	\scriptsize
	\caption{The pairwise kappa value of model verifiers for MNIST.}
	\vspace{-2mm}
	\label{fig:kappa-mnist}
	\begin{tabular}{|c|c|c|c|c|c|c|c|c|c|c|c|}
		\hline
		& \textbf{TM} & \textbf{$\text{V}_1$} & \textbf{$\text{V}_2$} & \textbf{$\text{V}_3$} & \textbf{$\text{V}_4$} & \textbf{$\text{V}_5$} & \textbf{$\text{V}_6$} & \textbf{$\text{V}_7$} & \textbf{$\text{V}_8$} & \textbf{$\text{V}_9$} & \textbf{$\text{V}_{10}$} \\ \hline
		\textbf{TM} & \cellcolor[HTML]{EFEFEF} & $0.79$ & $0.65$ & $0.78$ & $0.66$ & $0.37$ & $0.79$ & $0.78$ & $0.81$ & $0.58$ & $0.78$ \\ \hline
		\textbf{$\text{V}_1$} & \cellcolor[HTML]{EFEFEF} & \cellcolor[HTML]{EFEFEF} & $0.67$ & $0.81$ & $0.65$ & $0.47$ & $0.77$ & $0.77$ & $0.76$ & $0.69$ & $0.75$ \\ \hline
		\textbf{$\text{V}_2$} & \cellcolor[HTML]{EFEFEF} & \cellcolor[HTML]{EFEFEF} & \cellcolor[HTML]{EFEFEF} & $0.67$ & $0.50$ & $0.39$ & $0.62$ & $0.62$ & $0.62$ & $0.56$ & $0.60$ \\ \hline
		\textbf{$\text{V}_3$} & \cellcolor[HTML]{EFEFEF} & \cellcolor[HTML]{EFEFEF} & \cellcolor[HTML]{EFEFEF} & \cellcolor[HTML]{EFEFEF} & $0.64$ & $0.48$ & $0.76$ & $0.77$ & $0.75$ & $0.66$ & $0.74$ \\ \hline
		\textbf{$\text{V}_4$} & \cellcolor[HTML]{EFEFEF} & \cellcolor[HTML]{EFEFEF} & \cellcolor[HTML]{EFEFEF} & \cellcolor[HTML]{EFEFEF} & \cellcolor[HTML]{EFEFEF} & $0.25$ & $0.69$ & $0.71$ & $0.67$ & $0.47$ & $0.71$ \\ \hline
		\textbf{$\text{V}_5$} & \cellcolor[HTML]{EFEFEF} & \cellcolor[HTML]{EFEFEF} & \cellcolor[HTML]{EFEFEF} & \cellcolor[HTML]{EFEFEF} & \cellcolor[HTML]{EFEFEF} & \cellcolor[HTML]{EFEFEF} & $0.35$ & $0.36$ & $0.36$ & $0.43$ & $0.31$ \\ \hline
		\textbf{$\text{V}_6$} & \cellcolor[HTML]{EFEFEF} & \cellcolor[HTML]{EFEFEF} & \cellcolor[HTML]{EFEFEF} & \cellcolor[HTML]{EFEFEF} & \cellcolor[HTML]{EFEFEF} & \cellcolor[HTML]{EFEFEF} & \cellcolor[HTML]{EFEFEF} & $0.84$ & $0.83$ & $0.56$ & $0.82$ \\ \hline
		\textbf{$\text{V}_7$} & \cellcolor[HTML]{EFEFEF} & \cellcolor[HTML]{EFEFEF} & \cellcolor[HTML]{EFEFEF} & \cellcolor[HTML]{EFEFEF} & \cellcolor[HTML]{EFEFEF} & \cellcolor[HTML]{EFEFEF} & \cellcolor[HTML]{EFEFEF} & \cellcolor[HTML]{EFEFEF} & $0.86$ & $0.58$ & $0.84$ \\ \hline
		\textbf{$\text{V}_8$} & \cellcolor[HTML]{EFEFEF} & \cellcolor[HTML]{EFEFEF} & \cellcolor[HTML]{EFEFEF} & \cellcolor[HTML]{EFEFEF} & \cellcolor[HTML]{EFEFEF} & \cellcolor[HTML]{EFEFEF} & \cellcolor[HTML]{EFEFEF} & \cellcolor[HTML]{EFEFEF} & \cellcolor[HTML]{EFEFEF} & $0.54$ & $0.82$ \\ \hline
		\textbf{$\text{V}_9$} & \cellcolor[HTML]{EFEFEF} & \cellcolor[HTML]{EFEFEF} & \cellcolor[HTML]{EFEFEF} & \cellcolor[HTML]{EFEFEF} & \cellcolor[HTML]{EFEFEF} & \cellcolor[HTML]{EFEFEF} & \cellcolor[HTML]{EFEFEF} & \cellcolor[HTML]{EFEFEF} & \cellcolor[HTML]{EFEFEF} & \cellcolor[HTML]{EFEFEF} & $0.54$ \\ \hline
		\textbf{$\text{V}_{10}$} & \cellcolor[HTML]{EFEFEF} & \cellcolor[HTML]{EFEFEF} & \cellcolor[HTML]{EFEFEF} & \cellcolor[HTML]{EFEFEF} & \cellcolor[HTML]{EFEFEF} & \cellcolor[HTML]{EFEFEF} & \cellcolor[HTML]{EFEFEF} & \cellcolor[HTML]{EFEFEF} & \cellcolor[HTML]{EFEFEF} & \cellcolor[HTML]{EFEFEF} & \cellcolor[HTML]{EFEFEF} \\ \hline
	\end{tabular}
\vspace{-6mm}
\end{table}
\begin{table*}[b]
	\vspace{-2mm}
	\scriptsize
	\centerfloat
	\caption{Prediction accuracy of the target model on MNIST, ten base model verifiers, three different teamings for model verification ensemble and denoising-verification cross-layer ensemble.}
	\vspace{-2mm}
	\label{fig:exp-teaming-mnist}
	\begin{tabular}{|l|l|c|c|c|c|c|c|c|c|c|c|c|c|c|}
		\hline
		\multicolumn{2}{|c|}{\multirow{2}{*}{\textbf{Model}}} & \multirow{2}{*}{\textbf{\begin{tabular}[c]{@{}c@{}}Benign\\ Accuracy\end{tabular}}} & \textbf{FGSM} & \textbf{BIM} & \multicolumn{2}{c|}{\textbf{CW$_\infty$}} & \multicolumn{2}{c|}{\textbf{CW$_2$}} & \multicolumn{2}{c|}{\textbf{CW$_0$}} & \multicolumn{2}{c|}{\textbf{JSMA}} & \multirow{2}{*}{\textbf{Average}} & \multirow{2}{*}{\textbf{Std}}  \\ \cline{4-13}
		\multicolumn{2}{|l|}{} & & \multicolumn{2}{c|}{\textbf{UA}} & \textbf{ML} & \textbf{LL} & \textbf{ML} & \textbf{LL} & \textbf{ML} & \textbf{LL} & \textbf{ML} & \textbf{LL} &  & \\ \hline
		\multicolumn{2}{|l|}{TM: Target Model} & $0.9943$ & $0.54$ & $0.08$ & $0.00$ & $0.00$ & $0.00$ & $0.00$ & $0.00$ & $0.00$ & $0.07$ & $0.47$ & $0.12$ & $0.21$ \\ \hline
		\multicolumn{2}{|l|}{$\text{V}_1$: CNN-5} & $0.9919$ & $0.58$ & $0.27$ & $0.59$ & $0.43$ & $0.74$ & $0.78$ & $0.72$ & $0.72$ & $0.90$ & $0.81$ & $0.65$ & $0.19$ \\ \hline
		\multicolumn{2}{|l|}{$\text{V}_2$: CNN-4} & $0.9861$ & $0.65$ & $0.44$ & $0.85$ & $0.68$ & $0.85$ & $0.74$ & $0.60$ & $0.57$ & $0.76$ & $0.63$ & $0.68$ & $0.13$ \\ \hline
		\multicolumn{2}{|l|}{$\text{V}_3$: CNN-6} & $0.9917$ & $0.65$ & $0.37$ & $0.81$ & $0.71$ & $0.82$ & $0.89$ & $0.73$ & $0.65$ & $0.86$ & $0.73$ & $0.72$ & $0.15$ \\ \hline
		\multicolumn{2}{|l|}{$\text{V}_4$: LeNet} & $0.9880$ & $0.49$ & $0.44$ & $0.72$ & $0.31$ & $0.72$ & $0.63$ & $0.70$ & $0.57$ & $0.86$ & $0.74$ & $0.62$ & $0.17$ \\ \hline
		\multicolumn{2}{|l|}{$\text{V}_5$: MLP} & $0.9761$ & $0.51$ & $0.44$ & $0.74$ & $0.61$ & $0.88$ & $0.87$ & $0.83$ & $0.81$ & $0.85$ & $0.74$ & $0.73$ & $0.16$ \\ \hline
		\multicolumn{2}{|l|}{$\text{V}_6$: MobileNet} & $0.9934$ & $0.19$ & $0.18$ & $0.28$ & $0.16$ & $0.49$ & $0.36$ & $0.76$ & $0.55$ & $0.95$ & $0.87$ & $0.48$ & $0.30$ \\ \hline
		\multicolumn{2}{|l|}{$\text{V}_7$: MobileNet-v2} & $0.9939$ & $0.22$ & $0.23$ & $0.36$ & $0.11$ & $0.34$ & $0.14$ & $0.70$ & $0.44$ & $0.96$ & $0.86$ & $0.44$ & $0.30$ \\ \hline
		\multicolumn{2}{|l|}{$\text{V}_8$: PNASNet} & $0.9950$ & $0.15$ & $0.15$ & $0.19$ & $0.08$ & $0.20$ & $0.10$ & $0.68$ & $0.43$ & $0.94$ & $0.80$ & $0.37$ & $0.32$ \\ \hline
		\multicolumn{2}{|l|}{$\text{V}_9$: SVM} & $0.9832$ & $0.67$ & $0.72$ & $0.98$ & $0.92$ & $\mathbf{0.98}$ & $\mathbf{0.97}$ & $0.93$ & $0.86$ & $0.96$ & $0.80$ & $0.88$ & $0.11$ \\ \hline
		\multicolumn{2}{|l|}{$\text{V}_{10}$: VGG} & $0.9946$ & $0.40$ & $0.32$ & $0.59$ & $0.03$ & $0.49$ & $0.35$ & $0.62$ & $0.32$ & $0.89$ & $0.80$ & $0.48$ & $0.25$ \\ \hhline{|=|=|=|=|=|=|=|=|=|=|=|=|=|=|=|}
		\multirow{3}{*}{{\begin{tabular}[l]{@{}l@{}}Model Verification \\ Ensemble\end{tabular}}} & Rand = $\text{V}_{1}$,$\text{V}_{2}$,$\text{V}_{4}$,$\text{V}_{7}$,$\text{V}_{10}$ & $\mathbf{0.9951}$ & $0.61$ & $0.39$ & $0.78$ & $0.51$ & $0.76$ & $0.69$ & $0.79$ & $0.68$ & $0.94$ & $0.80$ & $0.70$ & $0.16$ \\ \cline{2-15}
		& Rand$\kappa$ = $\text{V}_{3}$,$\text{V}_{4}$,$\text{V}_{10}$ & $0.9919$ & $0.63$ & $0.42$ & $0.84$ & $0.65$ & $0.88$ & $0.88$ & $0.80$ & $0.77$ & $0.92$ & $0.80$ & $0.76$ & $0.15$ \\ \cline{2-15}
		& Best$\kappa$ = $\text{V}_{5}$,$\text{V}_{6}$,$\text{V}_{9}$ & $0.9888$ & $0.71$ & $0.78$ & $0.93$ & $0.85$ & $0.96$ & $0.94$ & $\mathbf{0.94}$ & $0.89$ & $\mathbf{0.98}$ & $0.88$ & $0.89$ & $0.09$ \\ \hhline{|=|=|=|=|=|=|=|=|=|=|=|=|=|=|=|}
		\multirow{3}{*}{{\begin{tabular}[l]{@{}l@{}}Denosing-Verification\\ Cross-Layer Ensemble\end{tabular}}} & Rand = $\text{V}_{1}$,$\text{V}_{2}$,$\text{V}_{4}$,$\text{V}_{7}$,$\text{V}_{10}$ & $0.9912$ & $\mathbf{0.90}$ & $0.76$ & $0.95$ & $\mathbf{0.98}$ & $0.86$ & $0.81$ & $0.83$ & $0.84$ & $0.95$ & $0.85$ & $0.87$ & $0.07$ \\ \cline{2-15}
		& Rand$\kappa$ = $\text{V}_{3}$,$\text{V}_{4}$,$\text{V}_{10}$ & $0.9873$ & $0.84$ & $0.71$ & $0.98$ & $0.94$ & $0.96$ & $0.92$ & $0.88$ & $0.88$ & $0.95$ & $0.88$ & $0.89$ & $0.08$ \\ \cline{2-15}
		& Best$\kappa$ = $\text{V}_{5}$,$\text{V}_{6}$,$\text{V}_{9}$ & $0.9855$ & $0.89$ & $\mathbf{0.86}$ & $\mathbf{0.99}$ & $0.97$ & $\mathbf{0.98}$ & $0.95$ & $\mathbf{0.94}$ & $\mathbf{0.92}$ & $\mathbf{0.98}$ & $\mathbf{0.90}$ & $\mathbf{0.94}$ & $\mathbf{0.04}$ \\ \hline
	\end{tabular}
\end{table*}
\begin{table*}[b]
	\vspace{-2mm}
	\scriptsize
	\centering
	\caption{Transferability of adversarial examples generated over the target model.}
	\vspace{-2mm}
	\label{fig:exp-transfer-mnist}
	\begin{tabular}{|c|c|c|c|c|c|c|c|c|c|c|c|c|c|c|}
		\hline
		\multicolumn{2}{|c|}{\textbf{}} & \textbf{TM} & \textbf{$\text{V}_1$} & \textbf{$\text{V}_2$} & \textbf{$\text{V}_3$} & \textbf{$\text{V}_4$} & \textbf{$\text{V}_5$} & \textbf{$\text{V}_6$} & \textbf{$\text{V}_7$} & \textbf{$\text{V}_8$} & \textbf{$\text{V}_9$} & \textbf{$\text{V}_{10}$} & \textbf{\begin{tabular}[c]{@{}c@{}}Model Verification\\ Ensemble\end{tabular}} & \textbf{\begin{tabular}[c]{@{}c@{}}Denoising-Verification\\Cross-Layer Ensemble\end{tabular}} \\ \hline
		FGSM & \multirow{2}{*}{UA} & $1.00$ & $0.72$ & $0.63$ & $0.74$ & $0.74$ & $0.67$ & $0.76$ & $0.76$ & $0.83$ & $0.57$ & $0.70$ & $0.63$ & $\mathbf{0.24}$ \\ \cline{1-1} \cline{3-15} 
		BIM &  & $1.00$ & $0.79$ & $0.61$ & $0.68$ & $0.61$ & $0.61$ & $0.83$ & $0.78$ & $0.86$ & $0.30$ & $0.71$ & $0.24$ & $\mathbf{0.15}$ \\ \hline
		\multirow{2}{*}{CW$_\infty$} & ML & $1.00$ & $0.34$ & $0.13$ & $0.19$ & $0.21$ & $0.14$ & $0.21$ & $0.35$ & $0.36$ & $0.02$ & $0.23$ & $0.06$ & $\mathbf{0.01}$ \\ \cline{2-15} 
		& LL & $1.00$ & $0.28$ & $0.06$ & $0.10$ & $0.15$ & $0.07$ & $0.04$ & $0.30$ & $0.16$ & $0.00$ & $0.15$ & $0.03$ & $\mathbf{0.00}$ \\ \hline
		\multirow{2}{*}{CW$_2$} & ML & $1.00$ & $0.24$ & $0.10$ & $0.18$ & $0.22$ & $0.08$ & $0.35$ & $0.55$ & $0.61$ & $0.02$ & $0.39$ & $0.04$ & $\mathbf{0.01}$ \\ \cline{2-15} 
		& LL & $1.00$ & $0.03$ & $0.04$ & $0.01$ & $0.06$ & $0.02$ & $0.07$ & $0.45$ & $0.45$ & $0.00$ & $0.21$ & $0.02$ & $\mathbf{0.00}$ \\ \hline
		\multirow{2}{*}{CW$_0$} & ML & $1.00$ & $0.25$ & $0.22$ & $0.24$ & $0.27$ & $0.11$ & $0.20$ & $0.20$ & $0.25$ & $0.03$ & $0.34$ & $0.05$ & $\mathbf{0.03}$ \\ \cline{2-15} 
		& LL & $1.00$ & $0.06$ & $0.12$ & $0.14$ & $0.06$ & $0.02$ & $0.11$ & $0.16$ & $0.19$ & $0.01$ & $0.16$ & $\mathbf{0.00}$ & $\mathbf{0.00}$ \\ \hline
		\multirow{2}{*}{JSMA} & ML & $1.00$ & $0.06$ & $0.13$ & $0.13$ & $0.10$ & $0.08$ & $0.05$ & $0.02$ & $0.02$ & $0.01$ & $0.08$ & $\mathbf{0.01}$ & $\mathbf{0.01}$ \\ \cline{2-15} 
		& LL & $1.00$ & $0.07$ & $0.19$ & $0.16$ & $0.05$ & $0.09$ & $0.00$ & $0.00$ & $0.00$ & $0.02$ & $0.00$ & $0.02$ & $\mathbf{0.00}$ \\ \hhline{|=|=|=|=|=|=|=|=|=|=|=|=|=|=|=|}
		\multicolumn{2}{|c|}{Average} & $1.00$ & $0.28$ & $0.22$ & $0.26$ & $0.25$ & $0.19$ & $0.26$ & $0.36$ & $0.37$ & $0.10$ & $0.30$ & $0.10$ & $\mathbf{0.05}$ \\ \hline
		\multicolumn{2}{|c|}{Std} & $0.00$ & $0.27$ & $0.22$ & $0.25$ & $0.24$ & $0.24$ & $0.30$ & $0.28$ & $0.31$ & $0.19$ & $0.24$ & $0.20$ & $\mathbf{0.08}$ \\ \hline
	\end{tabular}
	\vspace{-4mm}
\end{table*}

To understand the effectiveness of three different ensemble teaming options, we report in Table~\ref{fig:exp-teaming-mnist} the defensibility of the ensembles randomly sampled according to the above definitions. We make three observations. First, the target model suffers different levels of failure under various attacks. In comparison, each of the model verifiers consistently provides better robustness against most of the attacks than that of the target model. For instance, each verifier has higher DSRs under CW attacks, JSMA attacks, and BIM attacks and comparable performance under FGSM. Second, verification ensemble defense has better robustness compared to each of the individual verifiers. For each verification ensemble, its average DSR is higher than that of individual ensemble member. Third, using an ensemble team from the $\kappa$-ranked list performs better in many cases than random ensemble selection because the ensembles in the $\kappa$-ranked list have low kappa value and thus high model diversity with respect to failure-independence or error correlation. In addition to the above observations, the cross-layer ensembles significantly outperform their corresponding verification ensembles with the same team and performs much better than individual ensemble member. One example is the Rand$\kappa$ team with $\text{V}_3, \text{V}_4, \text{V}_{10}$ having an average DSR over all attacks of $0.72$, $0.62$, and $0.48$ respectively while the cross-layer ensemble method makes a significant improvement to $0.89$, demonstrating the complementary property of denoising and verification ensembles. 

The above shows that all three MODEF ensembles improve the test accuracy of the target model under eleven attacks on the two benchmark datasets. This indicates that ensembles with good diversity hold great potential for improving robustness of DNN models in the presence of adversarial inputs. We notice that verification ensemble offers a stronger ability to  repair adversarial examples with correct predictions while denoising ensemble tends to be more robust by flagging and rejecting adversarial examples. The cross-layer ensemble advocates high robustness with a high DSR by achieving a high PSR, which is in stark contrast to existing detection-only methods~\cite{xu2017feature}. 

To gain a better understanding of how the transferability of adversarial examples behaves under \scheme{}, we measure the attack transferability in ASR and report the results in Table~\ref{fig:exp-transfer-mnist}. Due to the space constraint, only the results on MNIST are presented. Given that the adversarial examples are generated over the target model (TM), the TM column shows the transferability to be $1.00$. Those adversarial examples may not be consistently transferable to other independently trained models. For instance, the adversarial examples generated by the CW$_2$ ML attack can be effectively transferred to $\text{V}_7$ and $\text{V}_8$ with a transferability of $0.55$ and $0.61$ respectively but achieve only a low transferability of $0.10$ for $\text{V}_2$, $0.08$ for $\text{V}_5$, and $0.02$ for $\text{V}_9$. Such divergence in transferability of adversarial examples is the main motivation of MODEF to exploit the weak spots of attack transferability using the denoising ensemble, the verification ensemble and the denoising-verification cross-layer ensemble as effective defense methods. As shown in the last two columns in Table~\ref{fig:exp-transfer-mnist}, the verification ensemble for MNIST has a very low transferability under all attacks. With the cross-layer ensemble that combines the denoising ensemble and verification ensemble, its attack transferability becomes further reduced.

\vspace{-1mm}
\section{Conclusion}\label{sec:conclusion}
We have presented \scheme{}, a diversity ensemble defense framework against adversarial deception in deep learning. Driven by quantifying ensemble diversity based on classification failure-independence, \scheme{} provides three diversity ensemble methods: (1) the model denoising ensemble, leveraging multiple diverse denoising autoencoders to remove adversarial perturbations, (2) the model verification ensemble that exploits the weak spots of attack transferability to verify and repair prediction output of the target model, and (3) the denoising-verification cross-layer ensemble, which can guard input and output of the target model through intelligently combining input denoising ensemble and output verification ensemble. Furthermore, \scheme{} is by design attack-independent and can generalize over different  attacks. Due to the space constraint, we could not include additional experimental results on the effectiveness of \scheme{} under new attack algorithms, including PGD~\cite{madry2017towards}.

\vspace{-1mm}
\section*{Acknowledgement}
This work was a part of XDefense umbrella project for robust diversity ensemble against deception. This research is partially sponsored by NSF CISE SaTC (1564097) and SAVI/RCN (1402266, 1550379). Any opinions, findings, and conclusions or recommendations expressed in this material are those of the author(s) and do not necessarily reflect the views of the National Science Foundation or other funding agencies and companies mentioned above.

\vspace{-1mm}
\bibliographystyle{IEEEtran}
\bibliography{references}

\end{document}